\documentclass[10pt,journal,compsoc]{IEEEtran}
\ifCLASSOPTIONcompsoc
  \usepackage[nocompress]{cite}
\else
  \usepackage{cite}
\fi
\ifCLASSINFOpdf
\else
\fi

\usepackage{graphicx}
\usepackage{threeparttable}
\usepackage{multirow}
\usepackage{array}
\usepackage[tight]{subfigure}
\usepackage{caption}
\usepackage{subfig}
\usepackage{setspace}
\usepackage{amsmath}
\usepackage{amssymb}
\usepackage{etoolbox}
\usepackage{algorithm}
\usepackage{comment}
\usepackage[noend]{algpseudocode}

\newcolumntype{P}[1]{>{\centering\arraybackslash}p{#1}}
\newcolumntype{M}[1]{>{\centering\arraybackslash}m{#1}}
\begin{document}
	\setlength{\abovedisplayskip}{5pt}
	\setlength{\belowdisplayskip}{5pt}
	\setlength{\textfloatsep}{15pt}
\title{End-to-End Latent Fingerprint Search}
% lists the Biometrics Council journals use for "first footnote" author

\author{Kai Cao,~\IEEEmembership{Member,~IEEE,}
        Dinh-Luan Nguyen,~\IEEEmembership{Student Member,~IEEE,} Cori Tymoszek,~\IEEEmembership{Student Member,~IEEE,} and~Anil~K.~Jain,~\IEEEmembership{Fellow,~IEEE}% <-this % stops a space
\IEEEcompsocitemizethanks{
\IEEEcompsocthanksitem
Kai Cao, Dinh-Luan Nguyen, Cori Tymoszek and A.K. Jain are with the Dept. of Computer Science and Engineering, Michigan State University, East Lansing, MI 48824 U.S.A. \protect\\
E-mail: \{kaicao,jain\}@cse.msu.edu
}
\thanks{}
}


\markboth{Journal of \LaTeX\ Class Files,~Vol.~14, No.~8, August~2015}%
{Shell \MakeLowercase{\textit{et al.}}: Bare Demo of IEEEtran.cls for Biometrics Council Journals}

% Biometrics Council journal papers as those publications place the marks
% or like this to get the Biometrics Council new two part style.
%\hspace{\columnsep}\makebox[\columnwidth]{Published by the IEEE Biometrics Council\hfill}}
% column for its text to clear the IEEEpubid mark (Biometrics Council jorunal


% for Biometrics Council papers, we must declare the abstract and index terms
\IEEEtitleabstractindextext{%
\begin{abstract}
Latent fingerprints are one of the most important and widely used sources of evidence in law enforcement and forensic agencies. Yet the performance of the state-of-the-art latent recognition systems is far from satisfactory, and they often require manual markups to boost the latent search performance. Further, the COTS systems are proprietary and do not output the true comparison scores between a latent and reference prints to conduct quantitative evidential analysis. We present an end-to-end latent fingerprint search system, including automated region of interest (ROI) cropping,  latent image preprocessing, feature extraction, feature comparison , and outputs a candidate list.  Two separate minutiae extraction models provide complementary minutiae templates.  To compensate for the small number of minutiae in small area and poor quality latents, a virtual minutiae set is generated to construct a texture template.  A 96-dimensional descriptor is  extracted for each minutia from its neighborhood.  For computational efficiency, the descriptor  length for virtual minutiae  is further reduced to 16  using product quantization. Our end-to-end system is evaluated   on three latent databases: NIST SD27 (258 latents); MSP  (1,200 latents), WVU  (449 latents) and N2N (10,000 latents) against a background set of 100K rolled prints, which includes the true rolled mates of the latents with rank-1 retrieval rates of 65.7\%, 69.4\%,  65.5\%, and 7.6\% respectively.  A multi-core solution implemented on 24 cores obtains 1ms per latent to rolled comparison. 

\end{abstract}

}

\maketitle

\IEEEdisplaynontitleabstractindextext

% \fi
\IEEEpeerreviewmaketitle

\IEEEraisesectionheading{\section{Introduction}\label{sec:introduction}}

\IEEEPARstart{L}{atent} fingerprints\footnote{Latent fingerprints are also known as latents or fingermarks} are arguably the most important forensic evidence that has been in use since 1893 \cite{FingerprintHandbook}. Hence, it is not surprising that fingerprint evidence at crime scenes is often regarded as ironclad.  This effect is compounded by the depiction of fingerprint evidence in media  in solving high profile crimes. For example, in the 2008 film The Dark Knight\footnote{https://www.imdb.com/title/tt5281134/} a shattered bullet is found at a crime scene. The protagonists create a digital reconstruction of the bullet's fragments, upon which a good quality fingermark is found, unaffected by heat or friction from the firing of the gun, nor by the subsequent impact. A match is quickly found in a fingerprint database, and the suspect's identity is revealed!

%\begin{figure}
%	\begin{center}
%		\includegraphics[width=0.75\linewidth]{Figure/san_diego_case.png}
%	\end{center}
%	\caption{On the right is a latent print taken from the San Diego crime scene in 1972. On the left, the fingerprint in the IAFIS database matched to the crime scene print by a San Diego Police Department latent print examiner. Image retrieved from \cite{SanDiego}.}
%	\label{fig:introduction}
%\end{figure}

\begin{figure}[t]	
	\begin{center}
	%\includegraphics[width=0.4\linewidth]{texture.jpg}
		%\captionsetup[subfigure]%{labelformat=empty}
		\subfigure[][]{
			\includegraphics[width=0.465\linewidth]{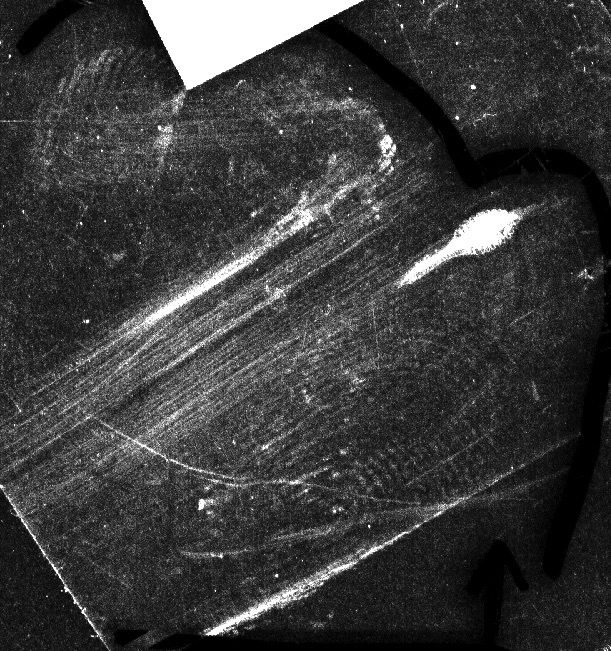}
		} \hspace{0.1cm}
		\subfigure[][]{
			\includegraphics[width=0.4\linewidth]{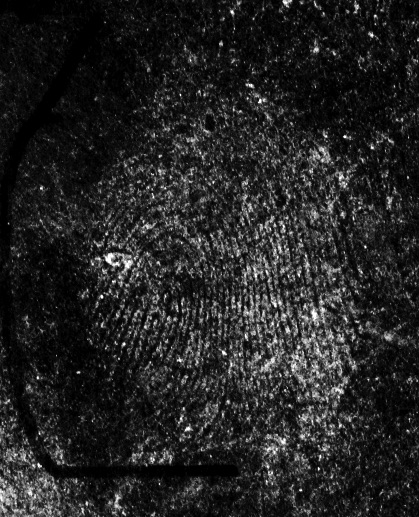}
		}\hspace{0.1cm}
		\subfigure[][]{
		\includegraphics[width=0.435\linewidth]{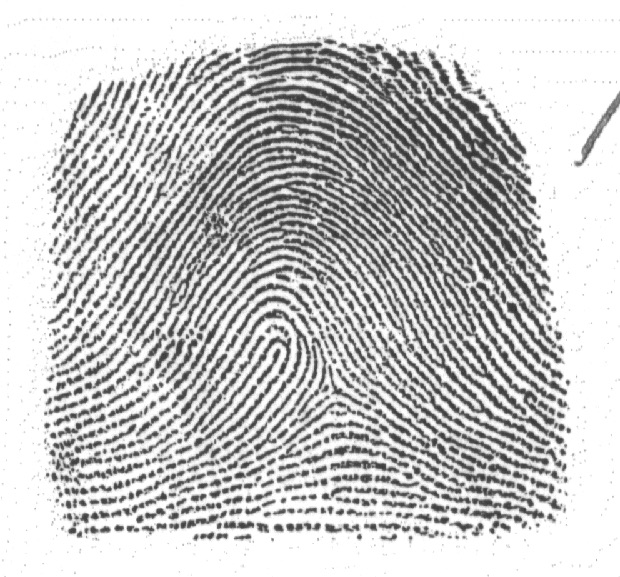}
		} \hspace{0.1cm}
		\subfigure[][]{
			\includegraphics[width=0.4\linewidth]{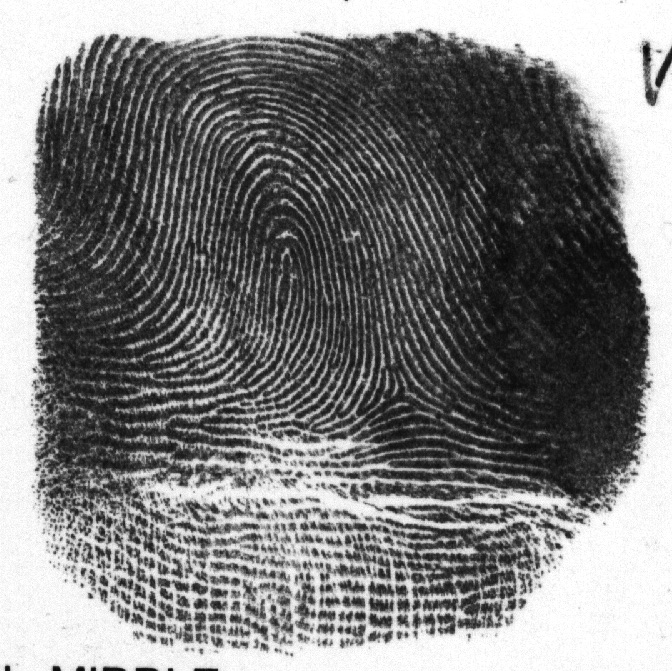}
		}\hspace{0.1cm}
	\end{center}
	\vspace{-0.4cm}
	\caption{Examples of low quality latents from the MSP latent database ((a) and (b)) and their true mates ((c) and (d)).} 	
	\label{fig:latent_examples}
\end{figure}

The above scenario, unfortunately, would likely have a much less satisfying outcome in the real forensic case work. While  processing of fingermarks has improved considerably due to advances in forensics, the problem of identifying latents, whether by forensic experts or automated systems, is far from solved. The primary difficulty in the analysis and identification of latent fingerprints is their poor quality  (See Fig. \ref{fig:latent_examples}). Compared to rolled and slap prints (also called reference prints or exemplar prints), which are acquired under supervision, latent prints are lifted after being unintentionally deposited by a subject, e.g., at crime scenes, typically resulting in poor quality in terms of ridge clarity and presence of large background noise. In essence, latent prints are partial prints, containing only a small section of the complete fingerprint ridge pattern. And unlike reference prints, investigators do not have the luxury of requesting a second impression from the culprit if the latent is found to be of extremely poor quality.

The significance of  research on latent identification is evident from the volume of latent fingerprints processed annually by publicly funded crime labs in the United States. A total of 270,000 latent prints were received by forensic labs for processing in 2009~\cite{crime_labs} which rose to 295,000 in 2014,  an increase of 9.2\%~\cite{crime_labs}. In June 2018, the FBI's Next Generation Identification (NGI) System received 19,766 requests for Latent Friction Ridge Feature Search (features need to be marked by an examiner) and 5,692 requests for Latent Friction Ridge Image Search (features are automatically extracted by IAFIS)~\cite{fbi_ngi}. These numbers represent an increase of 6.8\% and 25.8\%, respectively, over June 2017~\cite{fbi_ngi}.  Every year, the Criminal Justice Information Services (CJIS) Division gives its \emph{Latent Hit of the Year Award} to latent print examiners and/or law enforcement officers who solve a major violent crime using the Bureau's Integrated Automated Fingerprint Identification System, or IAFIS\footnote{https://www.fbi.gov/video-repository/newss-latent-hit-of-the-year-program-overview/view.}. %The 2010 FBI Latent Hit of the Year was awarded to San Diego Police Department  for  its role in identifying the killer in a 1972 cold case\footnote{https://archives.fbi.gov/archives/news/stories/2010/october/ latent\_102510/latent\_102510} by submitting the latent prints collected from the victim's  house to IAFIS. % as shown in Fig. \ref{fig:introduction}.

National Institute of Standards \& Technology (NIST) periodically conducts technology evaluations of fingerprint recognition algorithms, both for rolled (or slap) and latent prints. In NIST's most recent evaluation of rolled and slap prints, FpVTE 2012, the best performing AFIS achieved a false negative identification rate (FNIR) of 1.9\% for single index fingers, at a false positive identification rate (FPIR) of 0.1\% using 30,000 search subjects (10,000 subjects with mates and 20,000 subjects with no mates)~\cite{fpvte}. For latent prints, the most recent evaluation is the NIST ELFT-EFS where the best performing automated latent recognition system could only achieve a rank-1 identification rate of 67.2\% in searching 1,114 latents against a background containing 100,000 reference prints~\cite{fpvte}. The rank-1 identification rate of the best performing latent AFIS was improved from 67.2\% to 70.2\%\footnote{The best accuracy using both markup and image is 71.4\% @ rank-1.}~\cite{elft_efs} when feature markup by a latent expert was also input, in addition to the latent images, to the AFIS. This gap between reference  and latent fingerprint recognition capabilities is primarily due to the poor quality of friction ridges in latent prints (See Fig. \ref{fig:latent_examples}). This underscores the need for developing automated latent recognition with both high speed and accuracy\footnote{Automated latent recognition is also referred to as \textit{lights-out recognition};  objective is to minimize the role of latent examiners in latent recognition.}.  An automated latent recognition system will also assist in developing quantitative assessment of validity and reliability measures\footnote{Commercial AFIS neither provide extracted latent features nor the true comparison scores.  Instead, only truncated and/or modified scores are reported.} for latent fingerprint evidence as highlighted in the 2016 PCAST \cite{PCAST} and  the 2009 NRC \cite{PathForward} reports.

%The 2010 FBI Latent Hit of the Year\footnote{FBI Latent Hit of the Year is awarded annually to an outstanding latent examiner or officer who solved a major violent crime by using the FBI's IAFIS database, https://www.fbi.gov/video-repository/newss-latent-hit-of-the-year-program-overview/view.} recognized the 1972 San Diego case, which was reopened in 2008 by the San Diego Police Department. Latent prints collected from the victim?s house back in 1972 were submitted to IAFIS. The system came up with 20 possible matches. A San Diego Police Department latent print expert compared the matches with the crime scene latents and made an identification?an individual who had been previously been tried and acquitted on murder charges in Texas. 

% The main modules of a latent AFIS include preprocessing (ROI segmentation, ridge flow estimation and ridge enhancement), feature (minutiae and texture) extraction and comparison. 

 In the biometrics literature, the first paper on latent recognition was published by Jain \textit{et al.}  \cite{Jain2008Latent} in 2008 by using manually marked minutiae, region of interest (ROI) and ridge flow. Later, Jain and Feng  \cite{Feng2011PAMI}  improved the identification accuracy by using  manually marked extended latent features, including ROI, minutiae, ridge flow, ridge spacing and skeleton. However, marking these extended features in poor quality latents is very time-consuming and might not be feasible. Hence, the follow-up studies focused on increasing the degree of automation, i.e., reduction in the numbers of  manually marked features for matching, for example, automated ROI cropping \cite{Choi2012, Zhang2013TIFS, CaoPAMI2014,Nguyen2018BTAS}, ridge flow estimation \cite{CaoICB2015, CaoPAMI2014, FengPAMI2014,Feng2013} and ridge enhancement \cite{LI2018SP, Prabhu2018, Vatsa2019}, deep learning based minutiae extraction \cite{Tang2016, Darlow2017IJCB, Tang2017ICB, Nguyen2018ICB}, and comparison \cite{Fierrez2014}.  However, these studies only focus on specific modules in a latent AFIS and  do not build an end-to-end system. 
 
 \begin{figure*}
	\begin{center}
		\includegraphics[width=0.85\linewidth]{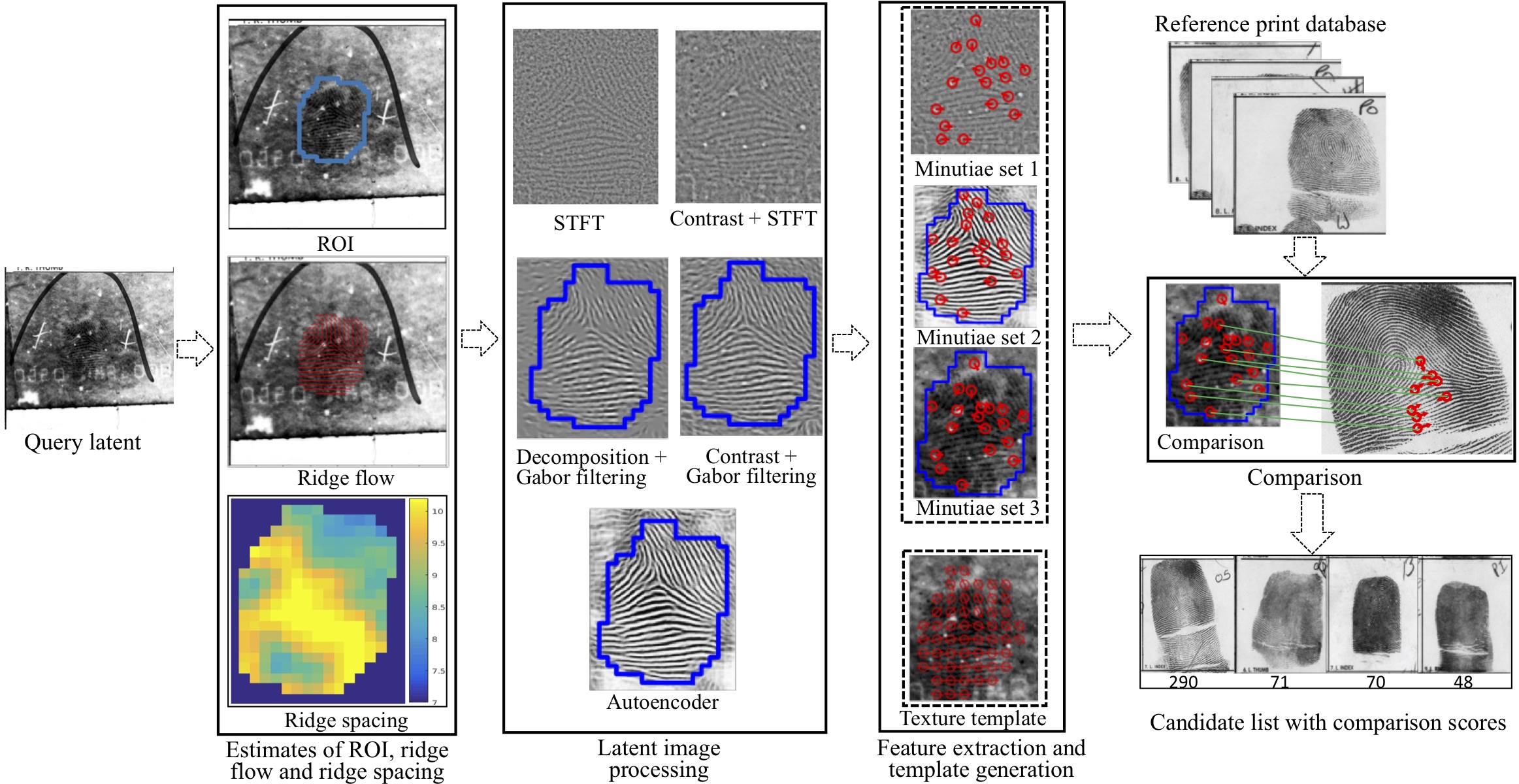}
	\end{center}
	\caption{Overview of the proposed end-to-end latent identification system. Given a query latent, three minutiae templates and one texture template are generated. Two matchers, i.e., minutiae template matcher and texture (virtual minutiae) template matcher are used for comparison between the query latent and reference prints.
		%The dotted arrow is an optional output , called the latent value, which is based on the quality and information content in the input latent.
	}
	\label{fig:flowchart}
\end{figure*}

Cao and Jain~\cite{Cao2018PAMI} proposed an automated latent recognition system which  includes automated steps of ridge flow and ridge spacing estimation, minutiae extraction, minutiae descriptor extraction, texture template (also called virtual minutiae template) generation and graph-based matching, and achieved the state-of-the-art accuracies on two latent databases, i.e., NIST SD27 and WVU latent databases. However, their study has the following limitations: (i) manually marked ROI is needed, (ii) skeleton-based minutiae extraction used in~\cite{Cao2018PAMI} introduces a large number of spurious minutiae, and (iii) a large texture template size (1.4MB) makes latent-to-reference comparison extremely slow.  Cao and Jain~\cite{Cao2018BTAS} improved both identification accuracy and search speed of texture templates by (i) reducing the template size, (ii)  efficient graph matching, and (iii) implementing the matching code in C++.  In this paper, we build a fully automated end-to-end system, and improve the search accuracy and computational efficiency of the system. We report results on three different latent fingerprint databases, i.e., NIST SD27, MSP and WVU, against a 100K background of reference prints.   
% \footnote{Only N2N is available in public domain.}

\section{Contributions}
The design and prototype of the proposed latent fingerprint search system is a substantially improved version of the work in~\cite{Cao2018PAMI}.  Fig. \ref{fig:flowchart} shows the overall flowchart of the proposed system. The main contributions of this paper are as follows:
 %  which typically includes  ROI cropping, ridge flow estimation, ridge enhancement, minutiae extraction and latent-to-rolled comparison. 
 
  \begin{itemize}
\item An autoencoder based latent fingerprint enhancement for robust and accurate extraction of ROI, ridge flow and ridge spacing.
\item An autoencoder based latent minutiae detection.
\item Complementary templates: three minutiae templates and one texture template.  These templates were selected from a large set of candidate templates to achieve the best recognition accuracy. 
\item Reducing descriptor length of minutiae template and texture template using non-linear mapping~\cite{Gong2018}. Descriptor for reference texture template is further reduced using product quantization for computational efficiency.  %The total latent (reference) template size is ?? bytes (?? bytes)
% \item Competitive latent identification  performance by using three minutiae templates and one texture template.
\item Latent search results on NIST SD27, MSP, and WVU latent databases against a background of 100K rolled prints show the state-of-the-art performance.

\item  A multi-core solution implemented on Intel(R) Xeon(R) CPU E5-2680 v3@2.50GHz takes $\sim$1ms per latent-to-reference comparison. Hence, a latent search against 100K reference prints can be completed in  100 seconds. Latent feature extraction time is $\sim$15 seconds on a machine with Intel(R) i7-7780@4.00GHz (CPU) and GTX 1080 Ti (GPU).
\end{itemize}

\section{Latent Preprocessing}
\subsection{Latent Enhancement via Autoencoder}
\begin{figure}[h]	
	\centering
	\includegraphics[width=0.9\linewidth]{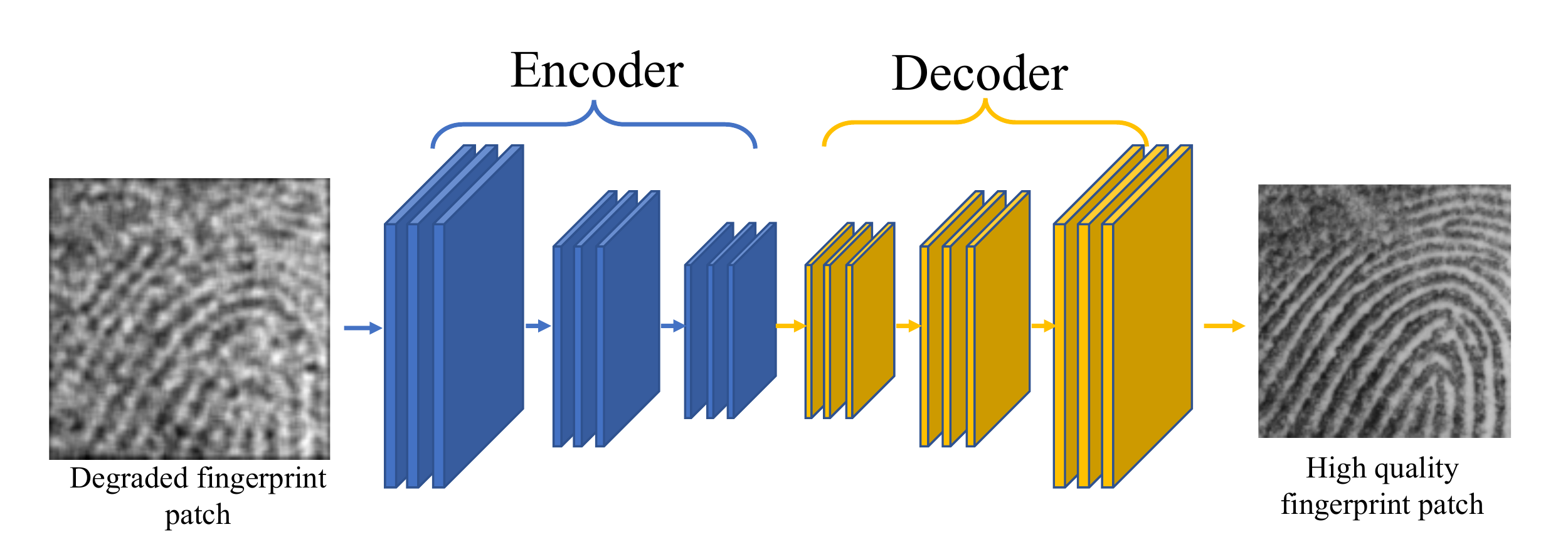}
			\caption{A convolutional autoencoder for latent enhancement. } 	
	\label{fig:enhancement_AEC}
\end{figure}

We present a convolutional autoencoder for latent enhancement. The enhanced images are required to find robust and accurate estimation of ridge quality, flow, and spacing. The flowchart for network training is shown in Fig. \ref{fig:enhancement_AEC}.    

\begin{figure}[h]	
	\begin{center}
	%\includegraphics[width=0.4\linewidth]{texture.jpg}
		%\captionsetup[subfigure]%{labelformat=empty}
			\subfigure[]{
			\includegraphics[width=0.15\linewidth]{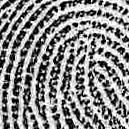}
		} \hspace{0.1cm}
		\subfigure[]{
			\includegraphics[width=0.15\linewidth]{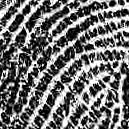}
		}\hspace{0.1cm}
		\subfigure[]{
			\includegraphics[width=0.15\linewidth]{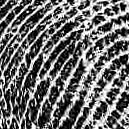}
		}\hspace{0.1cm}
		\subfigure[]{
			\includegraphics[width=0.15\linewidth]{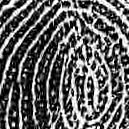}
		}\hspace{0.1cm}
		\subfigure[]{
			\includegraphics[width=0.15\linewidth]{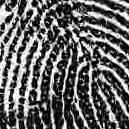}
		}\hspace{0.1cm}
		\subfigure[]{
			\includegraphics[width=0.15\linewidth]{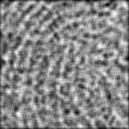}
		} \hspace{0.1cm}
		\subfigure[]{
			\includegraphics[width=0.15\linewidth]{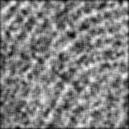}
		}\hspace{0.1cm}
		\subfigure[]{
			\includegraphics[width=0.15\linewidth]{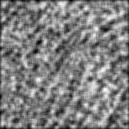}
		}\hspace{0.1cm}
		\subfigure[]{
			\includegraphics[width=0.15\linewidth]{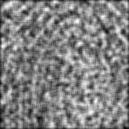}
		}\hspace{0.1cm}
		\subfigure[]{
			\includegraphics[width=0.15\linewidth]{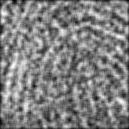}
		}
	\end{center}
	\vspace{-0.4cm}
	\caption{Fingerprint patch pairs (128$\times$ 128 pixels) consisting of high quality patches (top row) and their corresponding degraded patches (bottom row) for training the autoencoder.} 	
	\label{fig:training_patches}
\end{figure}

\begin{figure}[h]	
	\begin{center}
		%\captionsetup[subfigure]%{labelformat=empty}
			\subfigure[]{
			\includegraphics[width=0.40\linewidth]{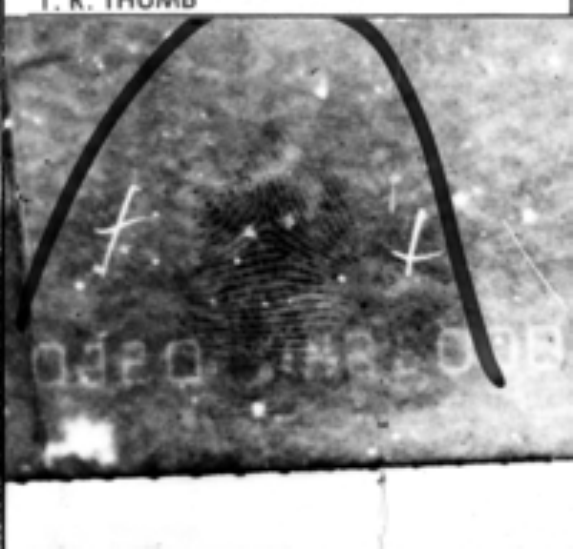}
		} \hspace{0.05cm}
		\subfigure[]{
			\includegraphics[width=0.40\linewidth]{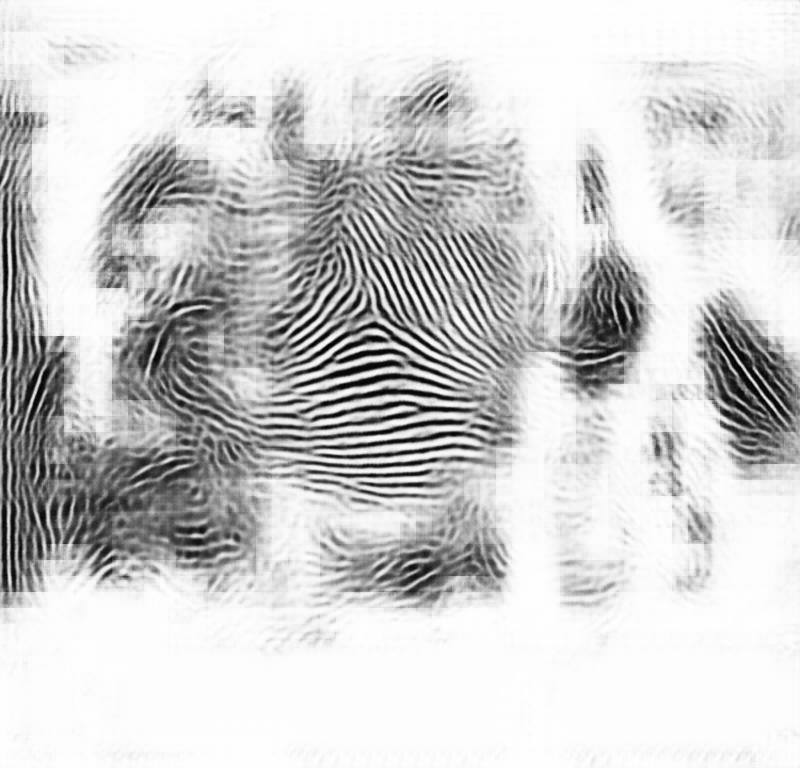}
		}\hspace{0.05cm}
			\subfigure[]{
			\includegraphics[width=0.40\linewidth]{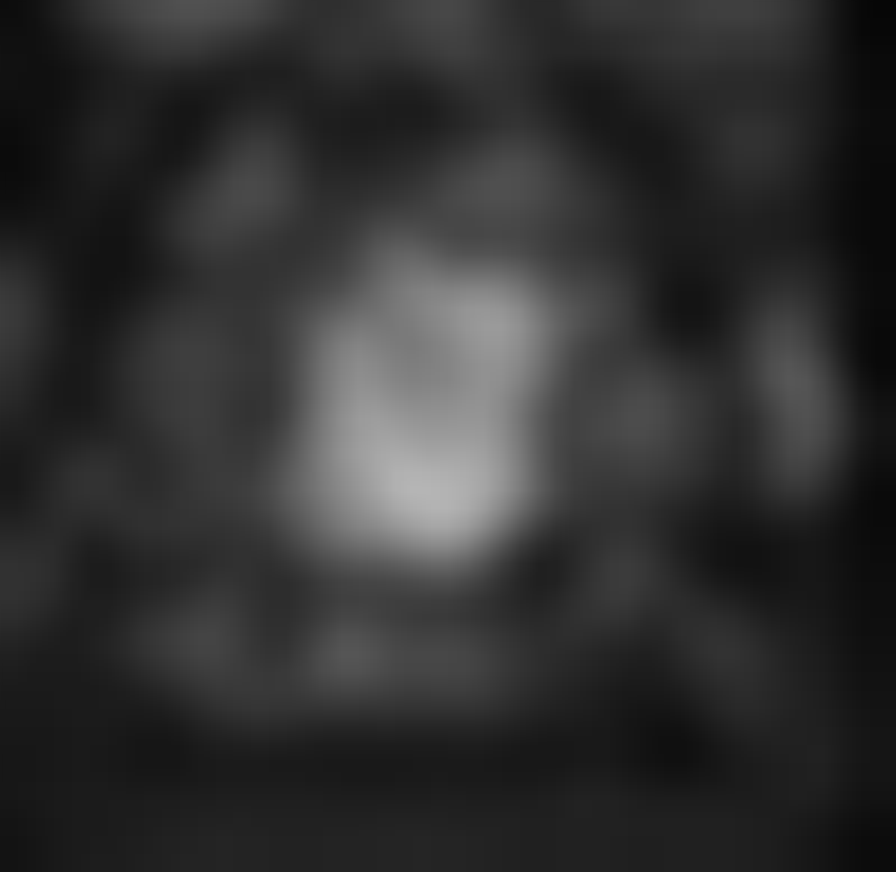}
		} \hspace{0.05cm}
		\subfigure[]{
			\includegraphics[width=0.40\linewidth]{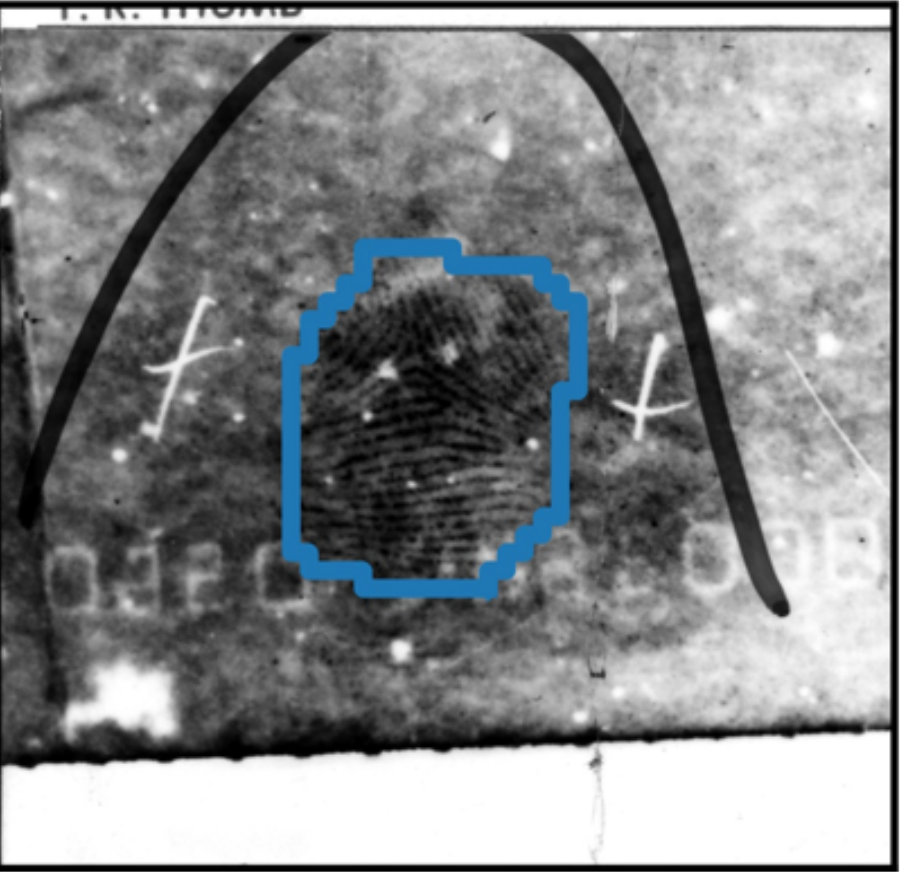}
		}
		\hspace{0.05cm}
		\subfigure[]{
			\includegraphics[trim={9.5cm 3cm 9.5cm 3cm},clip,width=0.40\linewidth]{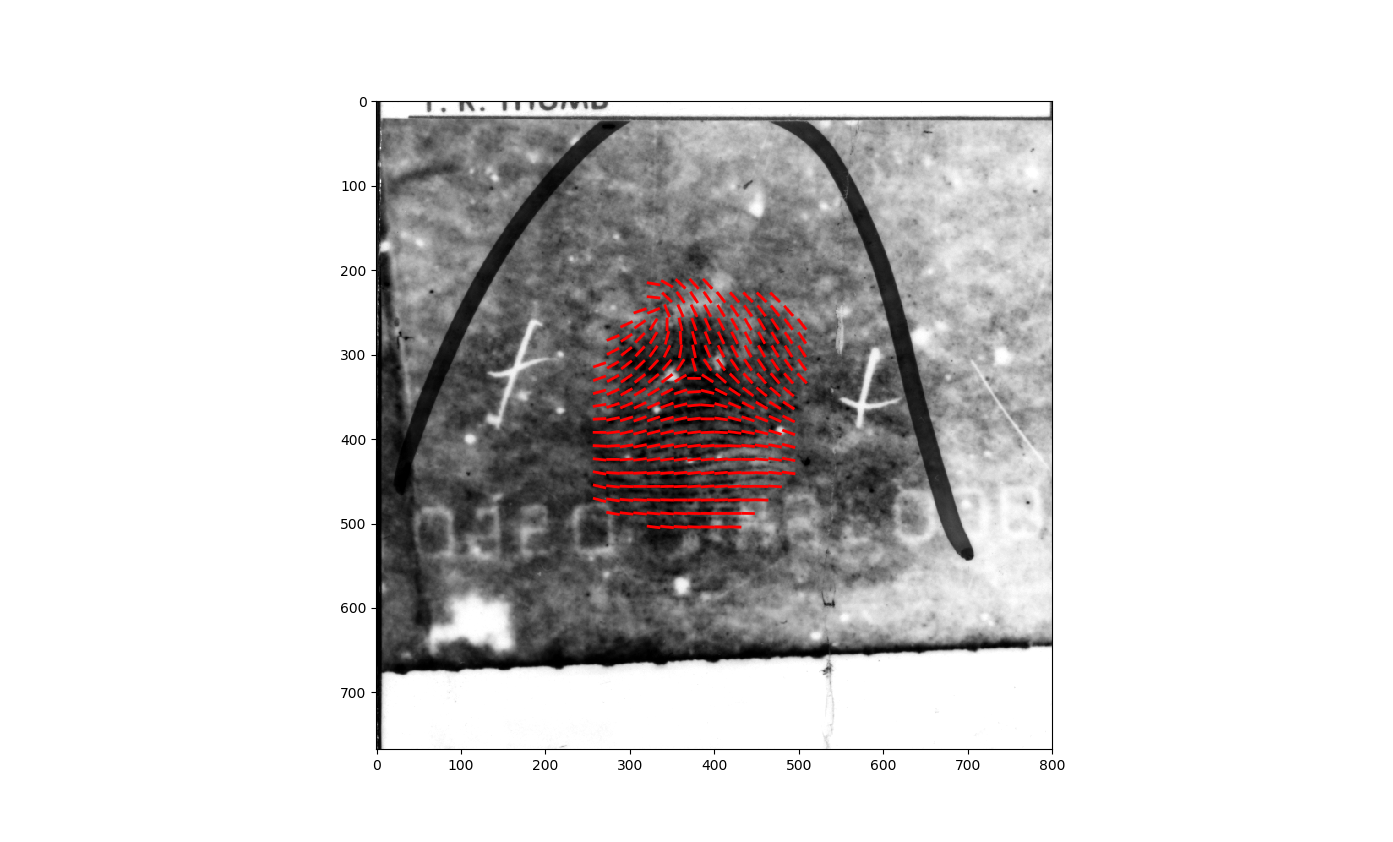}
			}
		\hspace{0.05cm}
		\subfigure[]{
			\includegraphics[width=0.38\linewidth]{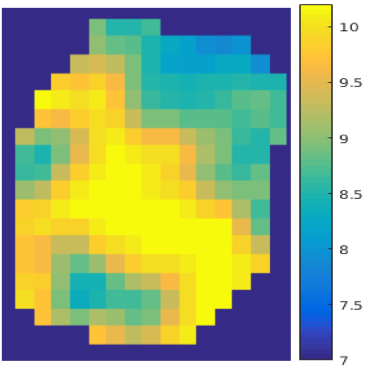}
			}	
	\end{center}
	\vspace{-0.4cm}
	\caption{Ridge quality, ridge flow and ridge spacing estimation.  (a) Latent fingerprint image,  (b)   enhanced using the autoencoder, (c) ridge quality estimated from (b) and ridge dictionary in Fig. \ref{fig:dictionary}, (d) cropping overlaid on the input latent image, (e) ridge flow overlaid on (the input latent image (a)) and (f) ridge spacing shown as a heat map } 	
	\label{fig:estimates}
\end{figure}

Since there is no publicly available dataset consisting of pairs of low quality and high quality fingerprint image for training the autoencoder, we degrade 2,000 high quality rolled fingerprint images (NFIQ 2.0\footnote{NFIQ 2.0 \cite{NFIQ2} ranges from 0 to 100, with 0 indicating the lowest quality and 100 indicating the highest quality fingerprint.} value $>$ 70)  to create image pairs for training. The degradation process involves randomly dividing fingerprint images into overlapping patches of size $128 \times 128$ pixels, followed by additive gaussian noise and Gaussian filtering with  a parameter $\sigma$ ($\sigma \in (5,15)$).   Fig. \ref{fig:training_patches} shows some examples of high quality fingerprint patches and their corresponding degraded versions. In addition, data augmentation methods (random rotation, random brightness and change in contrast) were used to improve the robustness of the trained autoencoder.

The convolutional autoencoder  includes an encoder and a decoder, as shown in Fig. \ref{fig:enhancement_AEC}. The encoder consists of 5 convolutional layers with a kernel size of $4\times 4$ and stride size of 2, while the decoder consists of 5 deconvolutional layers (or  transposed convolutional layer \cite{CNNguide}) also with a kernel size of $4\times 4$ and stride size of 2. The activation function ReLU (Rectified Linear Units) is used after each convolutional layer or deconvolutional layer with the exception of the last output layer, where the \textit{tanh} function is used. Table \ref{tab:Architecture} summarizes the architecture of the  convolutional Autoencoder. %Although the model is trained using patches of size   $128 \times 128$ pixels, it can be applied to any fingerprint image larger than  $128 \times 128$ pixels due to the nature of fully convolutional neural networks.

\begin{table}
	\caption{The network architecture of  autoencoder.  Size In and Size Out columns follow the format
		of $height\times width \times \#channels$.  Kernel column
		follows the format of $height\times width$, $stride$. Conv and Deconv denote convolutional layer and deconvolutional layer (or  transposed convolutional layer), respectively.}
	\small
	\begin{center}
		\begin{tabular}{|p{1.5cm}|p{2.cm}|p{1.8cm}|p{1.2cm}|}
			%{\linewidth}{M{0.14\linewidth}|M{0.1\linewidth}|M{0.1\linewidth}|M{0.1\linewidth}|M{0.1\linewidth}|M{0.1\linewidth}}
			\hline 
			\textbf{Layer}               &  \textbf{Size In} & \textbf{Size Out} &  \textbf{Kernel}\\
			\hline
			Input & 128$\times$128 $\times$ 1 & - & - \\
			\hline
			\hline
			Conv1  & 128$\times$128 $\times$ 1  & 64$\times$64 $\times$16  &4$\times$4, 2\\
			\hline
			Conv2  &  64$\times$64 $\times$16  & 32$\times$32 $\times$32 &4$\times$4, 2\\
			\hline
			Conv3  &  32$\times$32 $\times$32 & 16$\times$16 $\times$64 &4$\times$4, 2\\
			\hline
			Conv4  & 16$\times$16 $\times$64 &8$\times$8 $\times$128 &4$\times$4, 2\\
			\hline
			Conv5  & 8$\times$8 $\times$128 & 4$\times$4 $\times$256 &4$\times$4, 2\\
			\hline
			Deconv1  &  4$\times$4 $\times$256 & 8$\times$8 $\times$128 &4$\times$4, 2\\
			\hline
			Deconv2  &   8$\times$8 $\times$128 & 16$\times$16 $\times$64 &4$\times$4, 2\\
			\hline
			Deconv3  &   16$\times$16 $\times$64 & 32$\times$32 $\times$32 &4$\times$4, 2\\
			\hline
			Deconv4  &   32$\times$32 $\times$32 & 64$\times$64 $\times$16 &4$\times$4, 2\\
			\hline
			Deconv5  &   64$\times$64 $\times$16 & 128$\times$128 $\times$1 &4$\times$4, 2\\
			\hline	
		\end{tabular}
	\end{center}
	\label{tab:Architecture}
\end{table} 

The autoencoder trained on rolled prints does not work very well in enhancing latent fingerprints. So, instead of raw latent images, we input only the texture component of the latent by image decomposition \cite{CaoPAMI2014} to the autoencoder. Fig. \ref{fig:estimates} (b) shows the enhanced latent corresponding to the latent image in Fig. \ref{fig:estimates} (a). The enhanced latents have significantly higher ridge clarity than input latent images. 

\subsection{Estimation of Ridge Quality, Ridge Flow and Ridge Spacing} 
\label{sec:estimates}

\begin{figure}[h]	
	\begin{center}
	\includegraphics[width=0.9\linewidth]{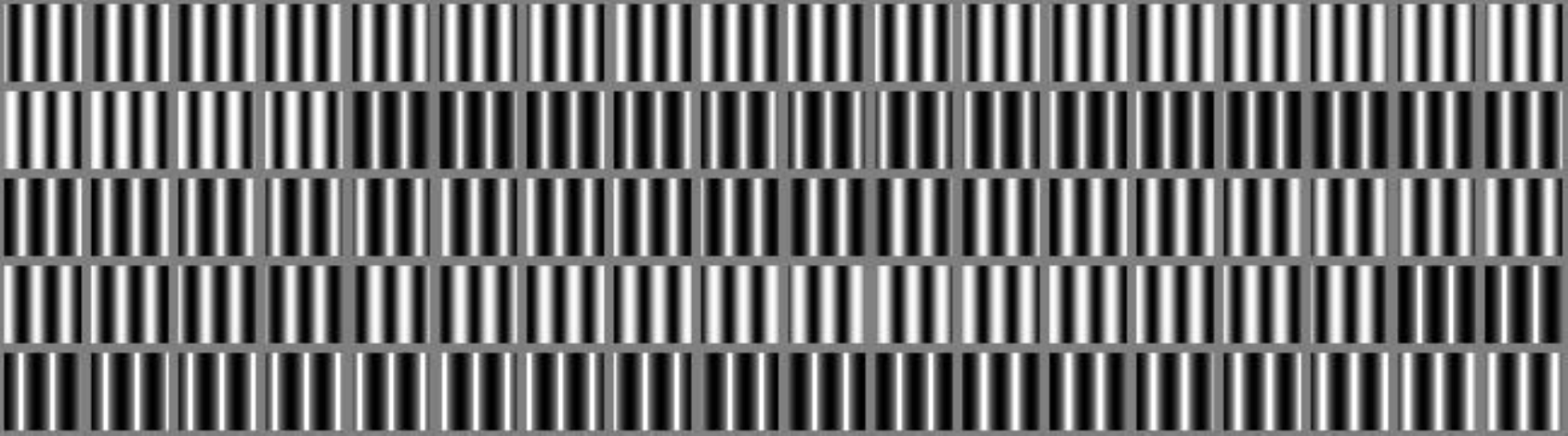}
	\end{center}
	\vspace{-0.4cm}
	\caption{Ridge structure dictionary (90 elements) for estimating ridge  quality, ridge flow and ridge spacing. The patch size of the dictionary elements is  $ 32\! \times \!32$ pixels. Figure retrieved from \cite{Cao2016ICB}.} 	
	\label{fig:dictionary}
\end{figure}

The dictionary based approach proposed in \cite{CaoPAMI2014} is modified as follows. Instead of learning the ridge structure dictionary using high quality fingerprint patches, we construct the dictionary elements with different ridge orientations and spacings using the approach described in \cite{Cao2016ICB}.  Fig. \ref{fig:dictionary} illustrates some of the dictionary elements in vertical orientation with different widths of ridges and valleys.

In order to estimate the ridge flow and ridge spacing, the enhanced latent image output by the autoencoder is divided into $32\times32$ patches with overlapping size of 16$\times$16 pixels. For each patch $P$, its similarity $s_i$ with each dictionary element $d_i$ (normalized to mean 0 and s.d. of 1) is computed as $s_i$ = $\frac{P\cdot d_i}{||P||+\alpha}$,  where $\cdot$ is the inner product, $||\cdot||$ denotes the $l_2$ norm and $\alpha$ ($\alpha=300$ in our experiments) is a regularization term.  The dictionary element $d_m$ with the maximum similarity $s_m$ ($s_m>=s_i, \forall i\neq m$) is selected and the ridge orientation and spacing of $P$ are regarded as the corresponding values of $d_m$.  The ridge quality of the patch $P_I$ in the input latent image  corresponding to $P$ is defined as the sum of $s_m$ and the similarity between $P_I$ and $P$. Figs. \ref{fig:estimates} (c), (d) and (f) show the ridge quality, ridge flow and ridge spacing, respectively.  Patches with ridge quality larger than $s_r$ ($s_r=0.35$ in our experiments) are considered as valid fingerprint patches. Morphological operations, including \emph{open} and \emph{close} operations, are used to obtain a smooth cropping.  Fig. \ref{fig:estimates} (d) shows the cropping (ROI) of the latent in Fig. \ref{fig:estimates} (a).

%So we propose three different levels of  enhancement for latent enhancement, i.e., i) contrast enhancement via  image decomposition \cite{CaoPAMI2014}, ii) STFT based enhancement \cite{STFT} which removes both high frequency and low frequency noise, and iii) Gabor filtering based enhancement \cite{Hong1998PAMI} which enhances ridge structures by using local ridge orientation and spacing. Fig. \ref{fig:preprocessing} shows three images created using different enhancement methods, where Gabor filtering based enhancement has better ridge clarity but also some artificial ridges around core point.  

\section{Minutiae Detection via Autoencoder}
A  convolutional Autoencoder-based minutiae detection approach is proposed in this section. Two minutiae extractor models are trained: one model (\emph{MinuNet\_reference}) is trained using manually edited minutiae on reference fingerprints while the other one (\emph{MinuNet\_Latent}) is fine-tuned based on \emph{MinuNet\_reference} using manually edited minutiae on latent fingerprint images.

\begin{figure}[t]	
	\centering
	\includegraphics[width=0.95\linewidth]{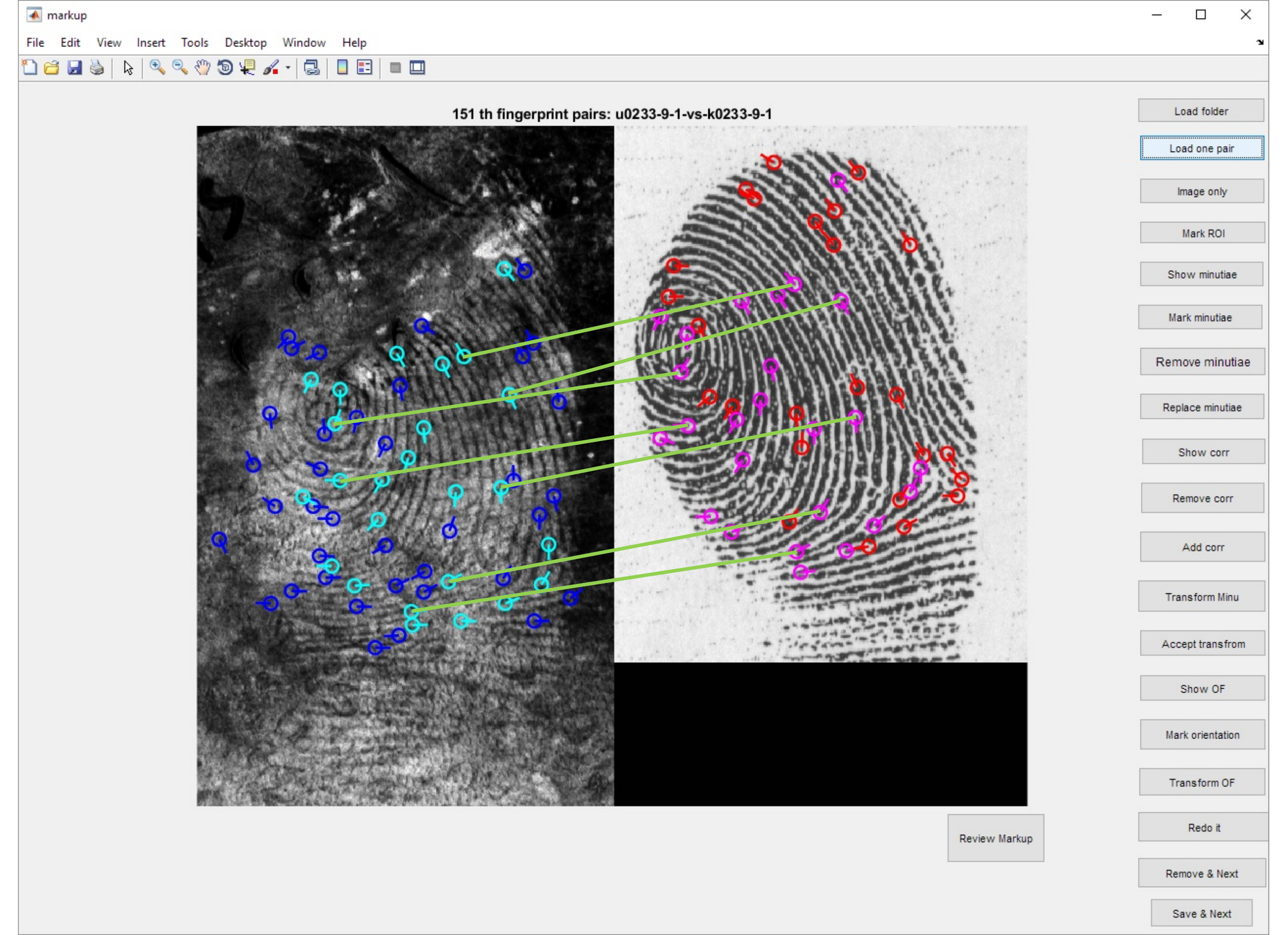}
			\caption{User interface for minutiae editing. It consists of operators for inserting, deleting, and repositioning minutia points. A latent and its corresponding rolled mate are illustrated here. Only the light blue minutiae in latent correspond with the pink minutiae in the mate shown in green lines.} 	
	\label{fig:tool}
\end{figure}
\subsection{Minutiae Editing}
In order to train networks for minutiae extraction for latent and reference fingerprints, a set of \emph{ground truth minutiae} are required. However, marking minutiae on poor quality latent fingerprint images and low quality reference fingerprint images is very challenging. It has been reported that even experienced latent examiners have low repeatability/reproducibility \cite{ulery2012repeatability} in minutiae markup. To obtain reliable minutiae ground truth, we designed a user interface to show a pair of latent and its corresponding reference fingerprint images side by side; the reference fingerprint image assists in editing minutiae on the latent.  The editing tool includes operations of insertion, deletion, and repositioning minutiae points (Fig. \ref{fig:tool}).   Instead of starting markup from scratch, some initial minutiae points and minutiae correspondences were generated using our automated minutiae detector and matcher. Because of this, we refer to this manual process  as \emph{minutiae editing} to distinguish it from markup from scratch.   
 
The following \emph{editing protocol}  was used on the initially marked minutiae points: i) remove spurious minutiae detected outside the ROI and those erroneously detected due to noise; ii) the locations of remaining minutiae points were adjusted as needed to ensure that they were accurately localized, iii) missing minutiae points which were visible in the image were marked; iv) minutiae correspondences between latent and its rolled mate were edited, including insertion and deletion; A thin plate spline (TPS) model was used to transform minutiae in latent and its rolled mate, and  v) a second round of minutiae editing (steps (i)-(iv) ) was conducted on latents. One of the authors carried out this editing process.

 % In this section, we compare several statistics as measured before and after editing, where "before editing" refers to the initial data automatically generated by the existing model. 

\begin{figure}[h]	
	\begin{center}
		%\captionsetup[subfigure]%{labelformat=empty}
			\subfigure[]{
			\includegraphics[width=0.45\linewidth]{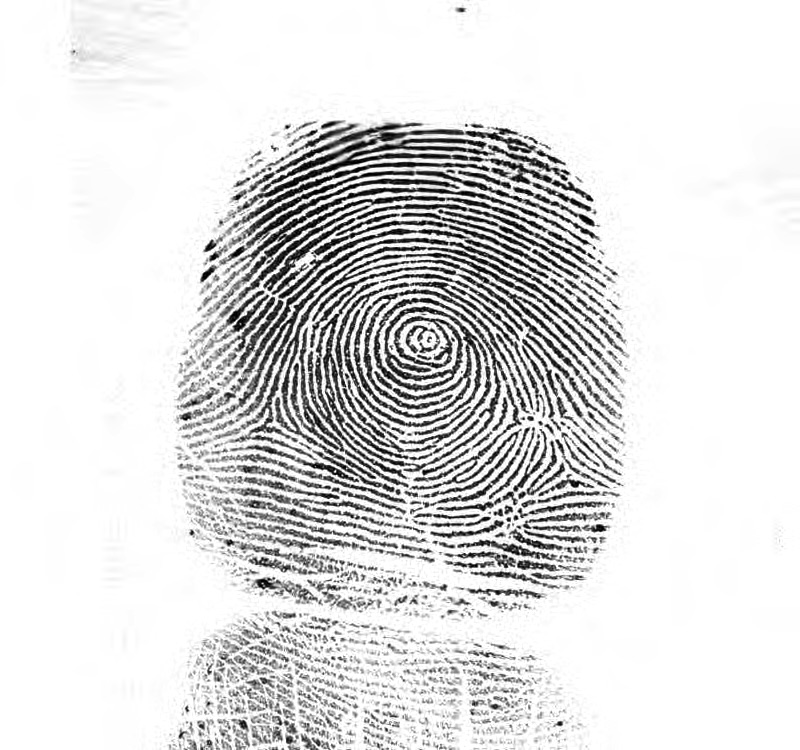}
		} 
		%\hspace{0.2cm}
		\subfigure[]{
			\includegraphics[width=0.45\linewidth]{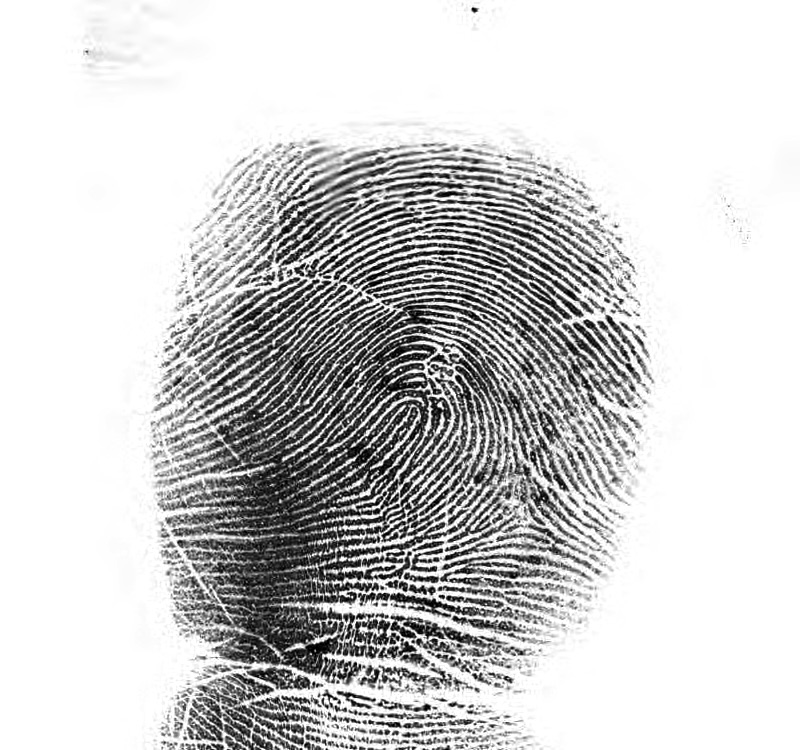}
		}
		%\hspace{0.2cm}
		\subfigure[]{
			\includegraphics[width=0.45\linewidth]{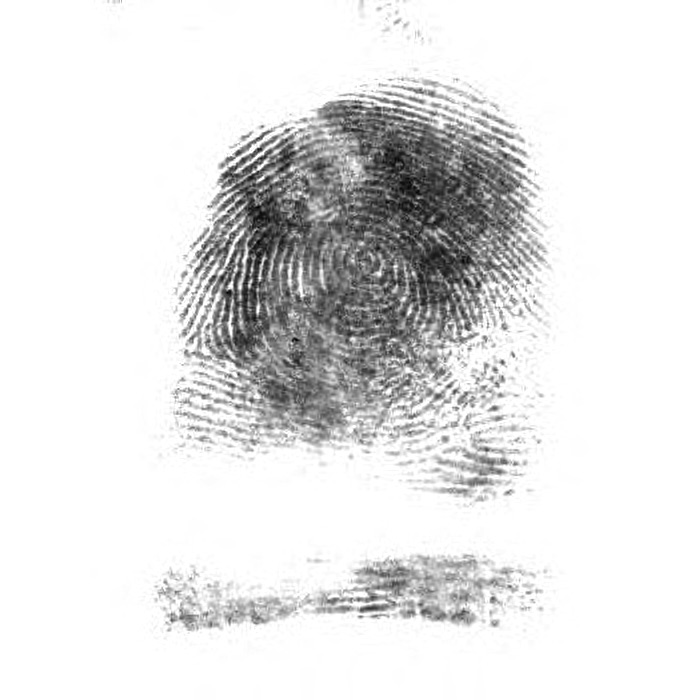}
		}
		%\hspace{0.2cm}
		\subfigure[]{
			\includegraphics[width=0.45\linewidth]{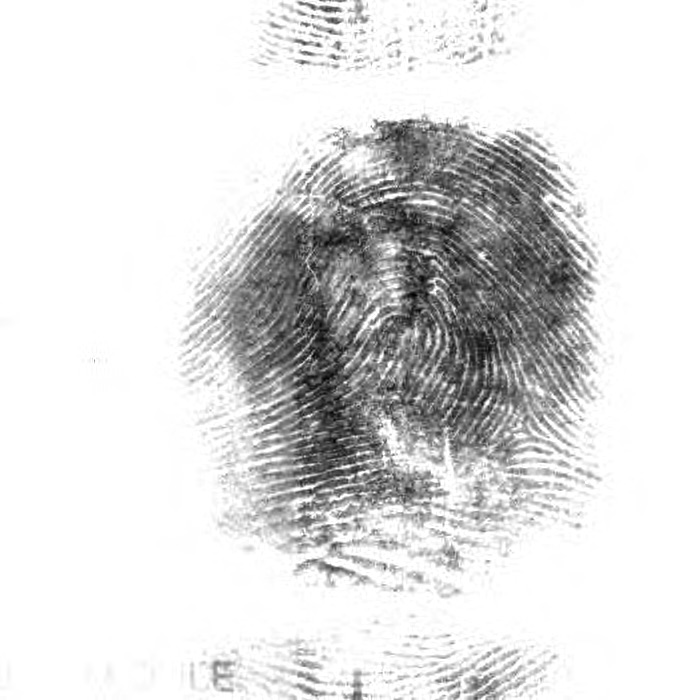}
		}
		%\hspace{0.2cm}
	\end{center}
	%\vspace{-0.4cm}
	\caption{Examples of rolled fingerprint images from the MSP  longitudinal database \cite{Yoon2015PNAS}  for training a network for minutiae detection for reference prints. The fingerprint images in the first row are of good quality while the corresponding fingerprint images of the same fingers in the second row are of low quality. The good quality fingerprint images in the first row were used to edit minutiae in poor quality fingerprint images in the second row. } 	
	\label{fig:training_eg}
\end{figure}

For training a minutiae detection model for reference fingerprints, i.e., \emph{MinuNet\_reference},  a total of 250 high quality and poor quality fingerprint pairs from 250 different fingers from the MSP longitudinal fingerprint database \cite{Yoon2015PNAS} were used.  A finger was selected if there is an impression (image) of it with the highest NFIQ 2.0 value $Q^h$ and the lowest NFIQ 2.0 value $Q^l$ which satisfies the following criterion ($Q^h - Q^l)> 70$. This ensured that we can obtain both high quality and low quality images for the same finger (See Fig. \ref{fig:training_eg}). A COTS SDK was used to get the initial minutiae and correspondences between selected fingerprint image pairs.

Given the significant differences in the characteristics of latents and rolled reference fingerprints, we fine-tuned the \emph{MinuNet\_reference}  model using minutiae in latent fingerprint images. A total of 300 latent and reference fingerprint pairs from the MSP latent database were used for retraining.  The minutiae detection model \emph{MinuNet\_reference} was used to extract initial minutiae points and a graph based minutiae matching algorithm proposed in \cite{Cao2018PAMI} was used to establish initial minutiae correspondences.

\subsection{Training Minutiae Detection Model}
\begin{figure}[h]	
	\centering
	\includegraphics[width=0.9\linewidth]{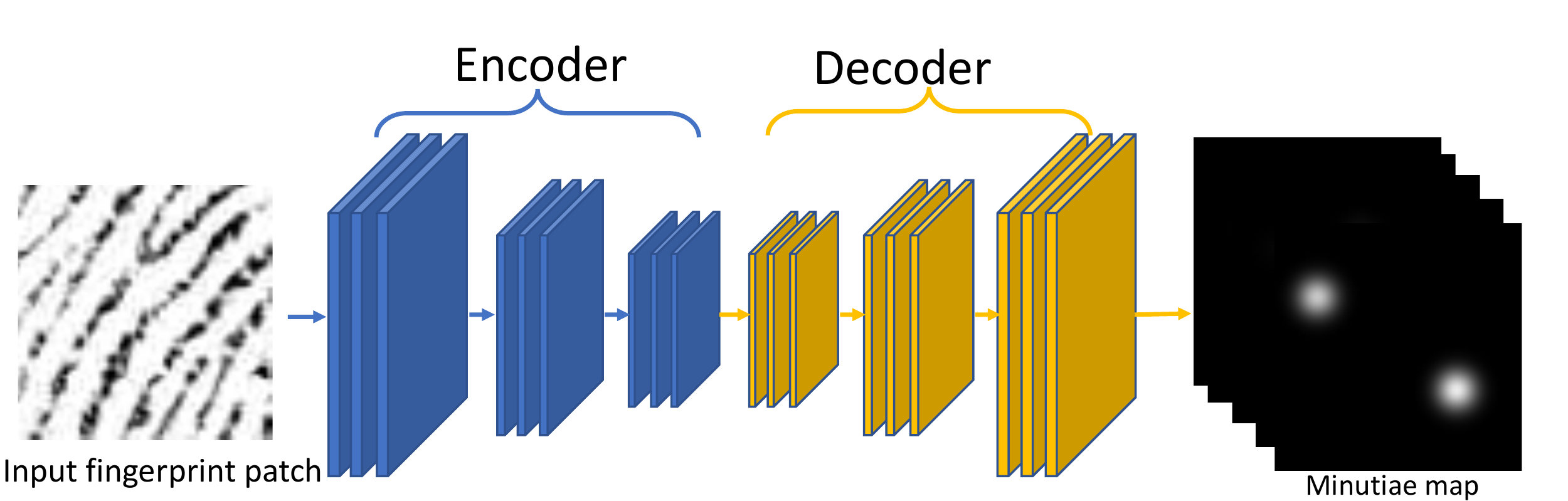}
			\caption{Training convolutional autoencoder for minutiae extraction.  For each input patch, the output is a 12-channel minutia map, where the $i$th channel represents the minutiae's contributions to orientation $i\cdot \pi/6$. } 	
	\label{fig:autoencoder}
\end{figure}

 Fig. \ref{fig:autoencoder} shows a  convolutional autoencoder-based network for minutiae detection. The advantages of this model include: i) a large training set since the image patches can be input to the network  instead of the whole images , and ii) generalization of the network to fingerprint images larger than the patches. In order to handle the variations in the number of minutiae in fingerprint patches, we encode the minutiae set as a 12-channel minutiae map and pose the training of minutiae detection model as a regression problem.  
 
\begin{figure}[h]	
	\centering
	\includegraphics[width=0.9\linewidth]{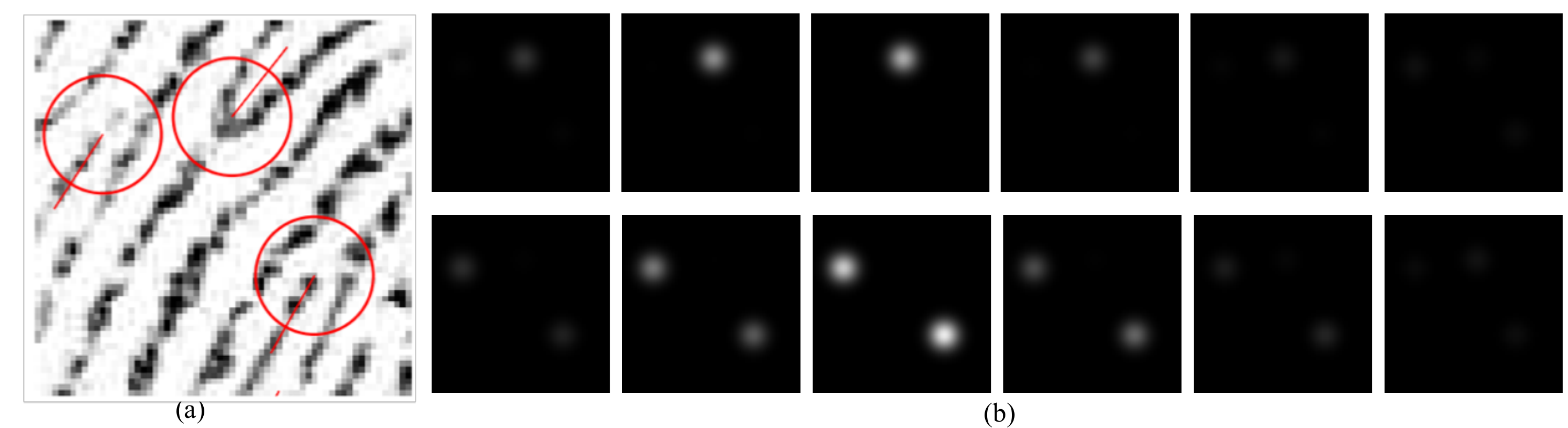}
			\caption{An example of a minutiae map. (a) Manually marked minutiae overlaid on a fingerprint patch and (b) 12-channel minutiae map. The  bright spots in the channel images indicate the location of minutiae points while the channel indicates the minutiae orientation. } 	
	\label{fig:minutiae_map}
\end{figure}

  A minutia point $m$ is typically represented as a triplet $m=(x,y,\theta)$, where $x$ and $y$ specify its location, and $\theta$ is its orientation (in the range $[0,2\pi]$). Inspired by minutia cylinder-code \cite{Cappelli2010PAMI}, we encode a minutiae set as a $c$-channel heat map and pose the minutiae extraction as a regression problem  ($c$=12 here). Let $h$ and $w$ be the height and width of the input fingerprint image $I$ and $T=\{m_1,m_2,..., m_n\}$ be its ISO/IEC 19794-2 minutiae template with $n$ minutiae points, where $m_t=(x_t,y_t,\theta_t)$, $t=1,...,n$. Its minutiae map $H\in R^{h\times w\times 12}$ is calculated by accumulating contributions from each minutiae point. Specifically, for each point $(i, j, k)$, a response value $M(i,j,k)$ calculated as 
\begin{equation}
\small
M(i,j,k) = \sum_{t=1}^{n} C_s((x_t,y_t),(i,j)) \cdot C_o(\theta_t,2k\pi/12)
\end{equation}
where the two terms $C_s((x_t,y_t),(i,j))$ and $C_o(\theta_t,2k\pi/12)$ are the spatial and orientation contributions of minutia $m_t$ to image point $(i, j, k)$,
respectively. $C_s((x_t,y_t),(i,j))$ is defined as a function of the
Euclidean distance between $(x_t,y_t)$ and $(i,j)$:
\begin{equation}
C_s((x_t,y_t),(i,j)) = exp(-\frac{||(x_t,y_t)-(i,j)||_2^2}{2\sigma_s^2}),
\end{equation}
where $\sigma_s$ is the parameter controlling the width of the Guassian. $C_o(\theta_t,2k\pi/12)$ is defined as a function of  the difference in orientation value between $\theta_t$ and $2k\pi/12$:
\begin{equation}
C_o(\theta_t,2k\pi/12) = exp(-\frac{d\phi(\theta_t,2k\pi/12)}{2\sigma_s^2}),
\end{equation}
and  $d\phi(\theta_1,\theta_2)$ is the orientation difference between angles $\theta_1$ and  $\theta_2$:
\begin{equation}
d\phi(\theta_1,\theta_2)= \begin{cases} |\theta_1 - \theta_2| &\text{ $-\pi \leq \theta_1-\theta_2<\pi$,}\\
2\pi-|\theta_1 - \theta_2| &\text{otherwise.}
\end{cases}
\end{equation}
Fig. \ref{fig:minutiae_map} illustrates 12-channel minutiae map, where the bright spots indicate the locations of minutiae points. This autoencoder architecture used for minutiae detection is similar to the autoencoder for latent enhancement with parameters specified in Table \ref{tab:Architecture}. The  three differences are thati i) the input fingerprint patches are size of $64\times 64$ pixels,  ii) the output is a 12-channel minutiae map rather than a single channel fingerprint image, and ii) the number of of convolutional layers and deconvolutional layers are 4 instead of 5. 

 %Data augmentation techniques, such as random cropping, random rotation and random brightness, were used for improving the robustness of the model. These fingerprint images are randomly divided into overlapping patches of size $128 \times 128$ pixels. These patches were degraded by first adding random gaussian noise, followed by the application of a Gaussian filter with random value of $\sigma$ ($\sigma \in (5,15)$).   Although the model was trained using patches of size $128 \times 128$ pixels, it can be applied to any fingerprint images larger than  $128 \times 128$ pixels due to the nature of fully convolutional neural networks.

The two minutiae detection models  introduced earlier, \emph{MinuNet\_reference} and \emph{MinuNet\_Latent}, are trained. For reference fingerprint images, the unprocessed fingerprint patches are used for training. On the other hand, latent fingerprint images were processed by short-time Fourier transform (STFT)  for training in order to alleviate the differences in latents; the model \emph{MinuNet\_Latent} is a fine-tuned version of the model \emph{MinuNet\_reference}.

\begin{figure}[h]	
	\centering
	\includegraphics[width=0.9\linewidth]{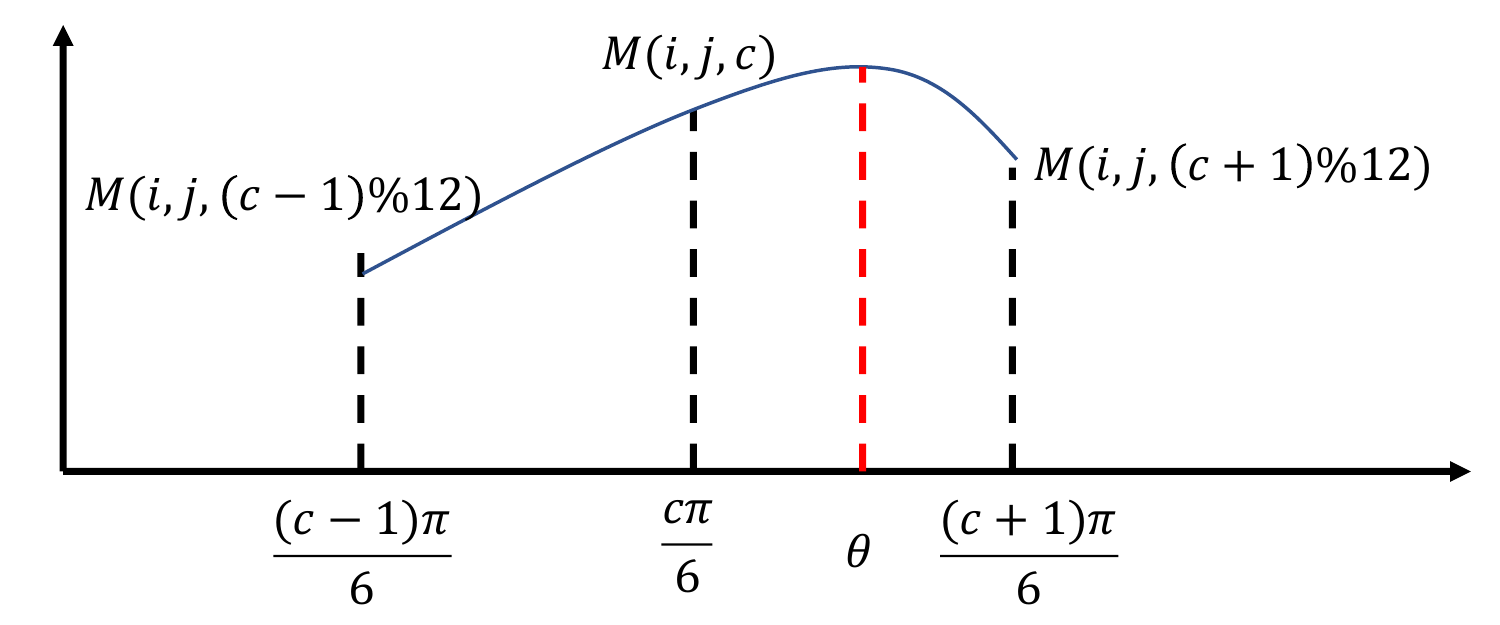}
			\caption{Minutia orientation ($\theta$) extraction using quadratic interpolation.} 	
	\label{fig:Interpolation}
\end{figure}

\begin{figure}[h]
	\centering
		\subfigure[]{
		\includegraphics[clip, trim=14cm 3cm 14cm 3cm, width=0.41\linewidth]{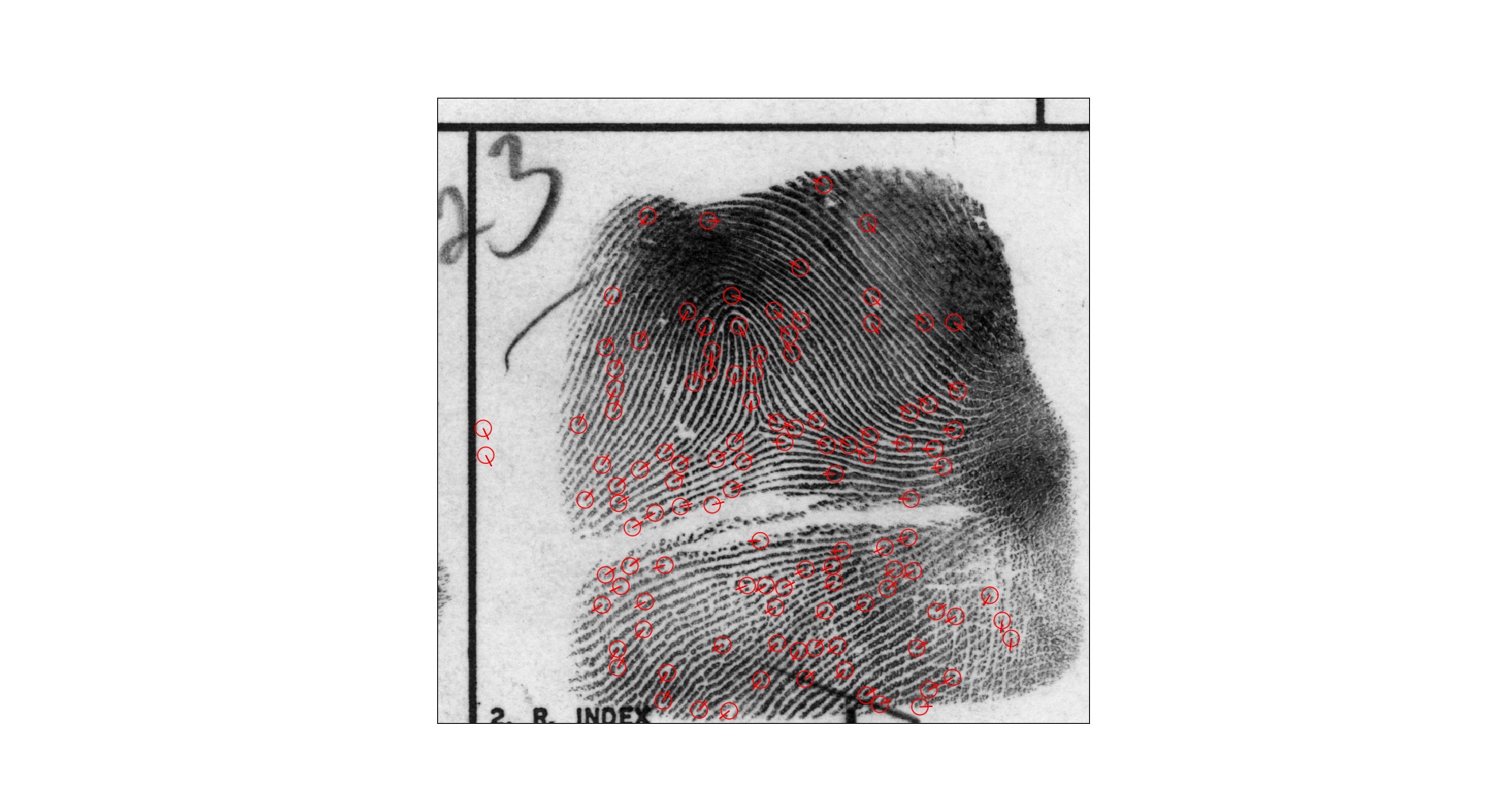}
	}  
	\subfigure[][]{
		\includegraphics[clip, trim=14cm 3cm 14cm 3cm, width=0.41\linewidth]{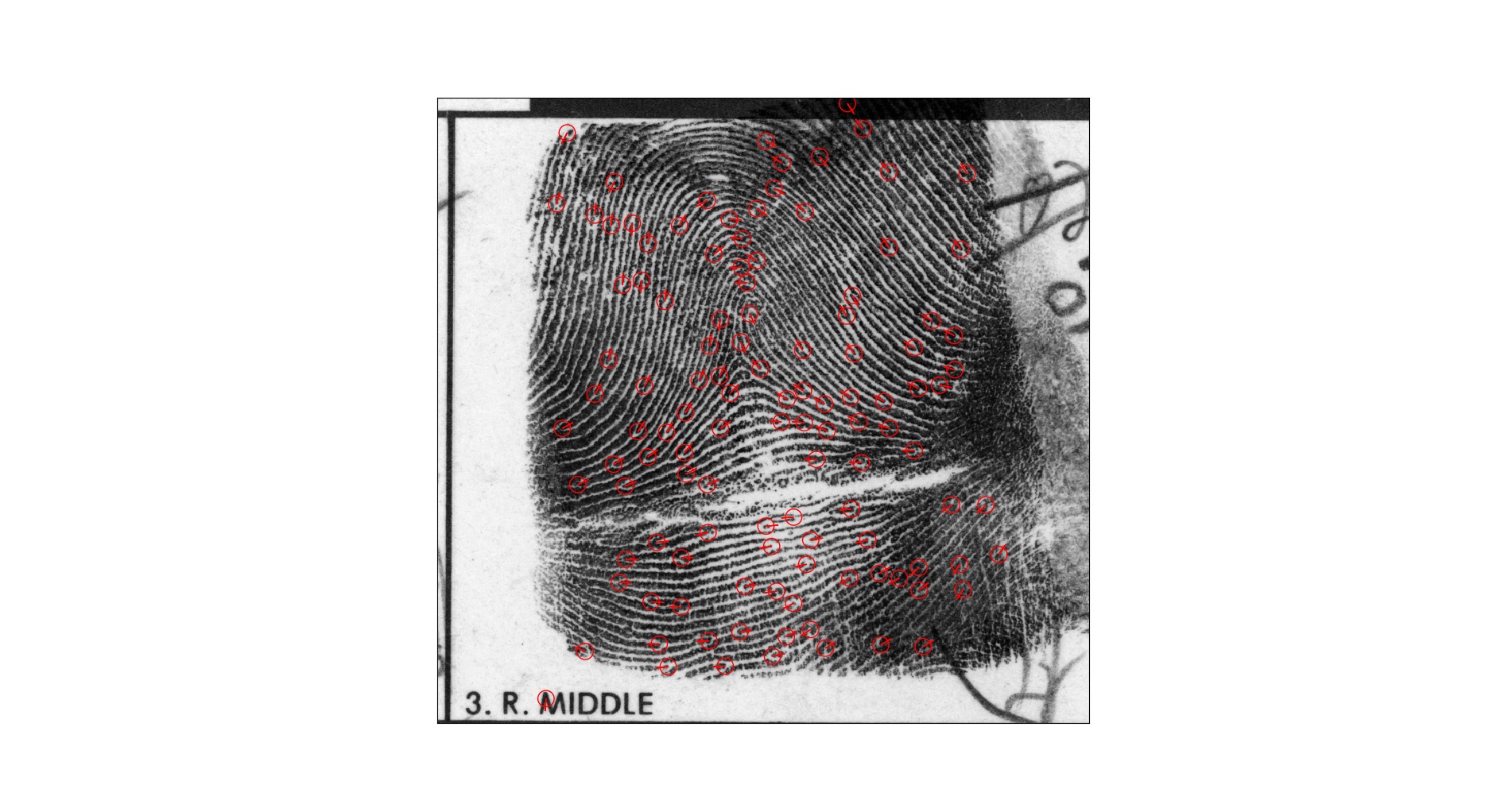}
	}   
	\subfigure[]{
		\includegraphics[clip, trim=14cm 3cm 14cm 3cm, width=0.41\linewidth]{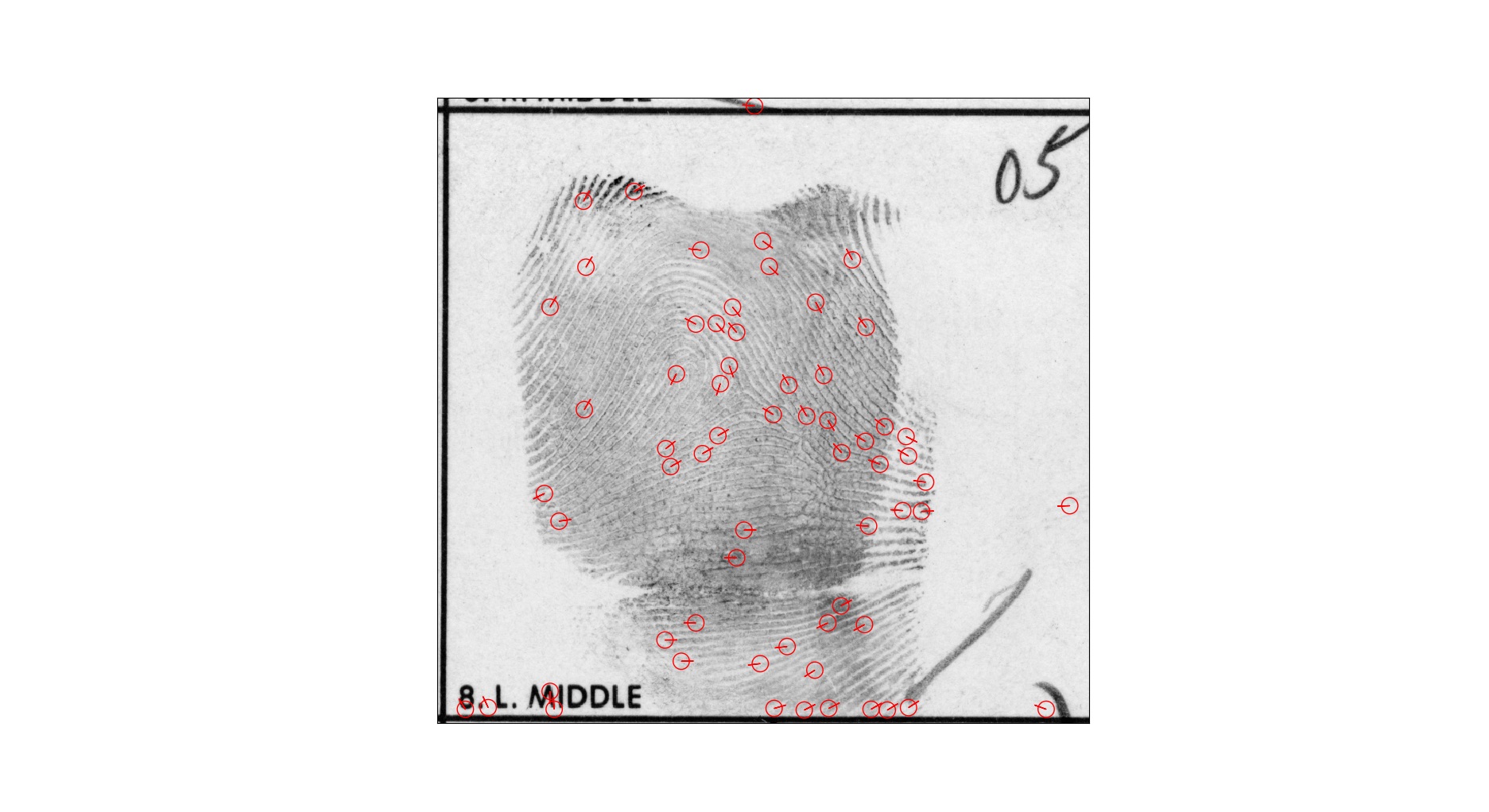}
	}  
	\subfigure[][]{
		\includegraphics[clip, trim=14cm 3cm 14cm 3cm, width=0.41\linewidth]{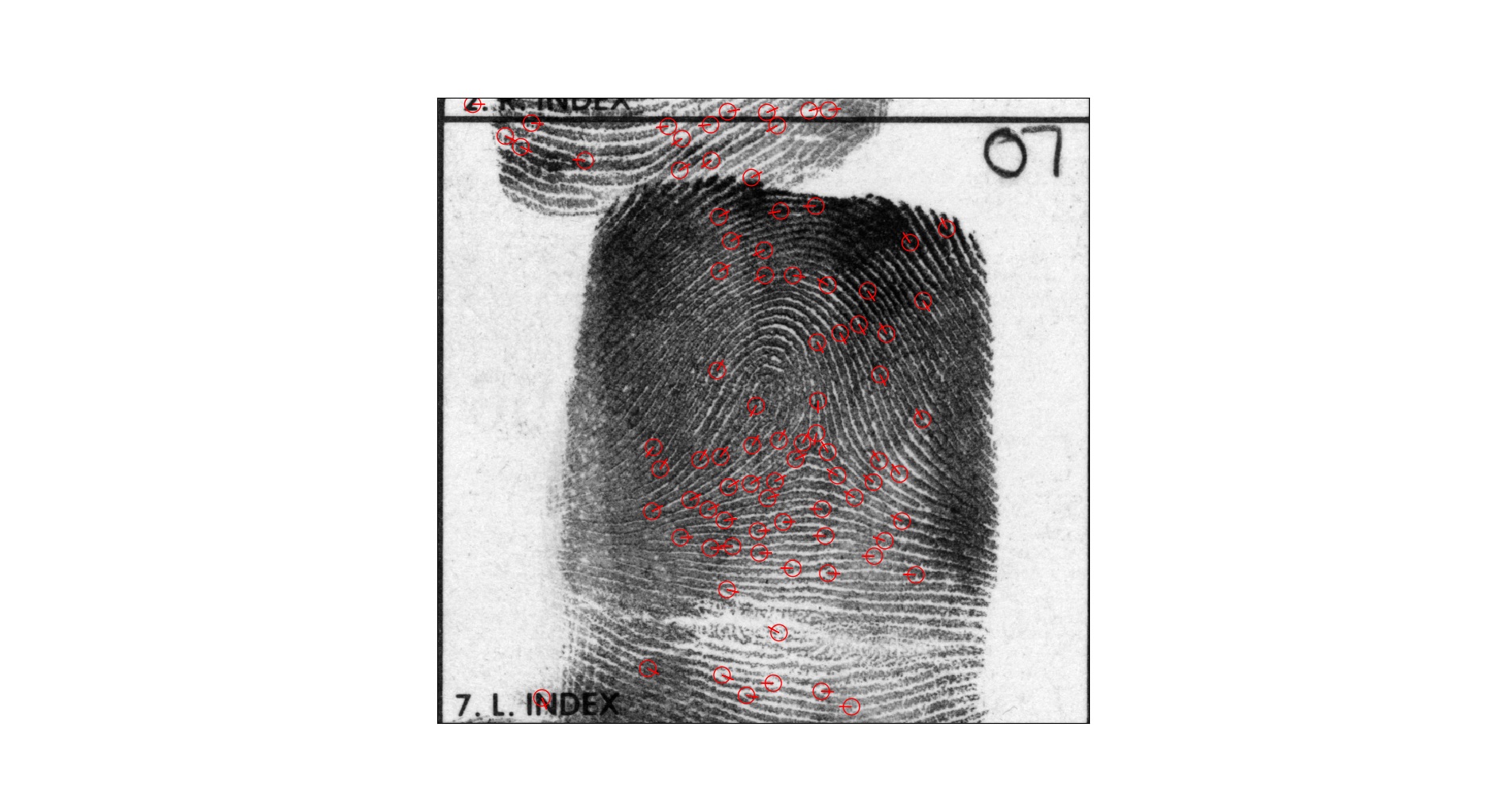}
	}
	  
	\caption{Examples of minutiae extracted on reference fingerprint images. Images in the first row are of good quality while the images in the second row are of poor quality.}
	\label{fig:minutiae_eg}
	\vspace*{-5pt}
\end{figure}

\subsection{Minutiae Extraction}

Given a fingerprint image of size $w\times h$ in the inference stage, a $w\times h \times 12$ minutiae map $M$ is output by a minutiae detection model.  For each location $(i,j,c)$ in $M$, if  $M(i,j,c)$ is larger than a threshold $m_t$ and it is a local max in its neighboring $5\times 5 \times 3$ cube,  a minutia is marked at location $(i,j)$. Minutia orientation $\theta$ is computed by maximizing the quadratic interpolation based on  $f({(c-1)\cdot \pi}/{6})=M(i,j,(c-1)\%12)$, $f( {c\cdot \pi}/{6}) = M(i, j, c)$ and $f( {(c+1)\cdot \pi}/{6})=M(i,j,(c+1)\%12)$, where $a\%b$ denotes $a$ modulo $b$. Fig. \ref{fig:Interpolation} illustrates minutia orientation estimation from the minutiae map.  Fig. \ref{fig:minutiae_eg} shows some examples of minutiae extracted in reference fingerprints.

\section{Minutia Descriptor} 
\label{sec:descriptor}

\begin{figure}[h]	
	\centering
	\includegraphics[width=0.9\linewidth]{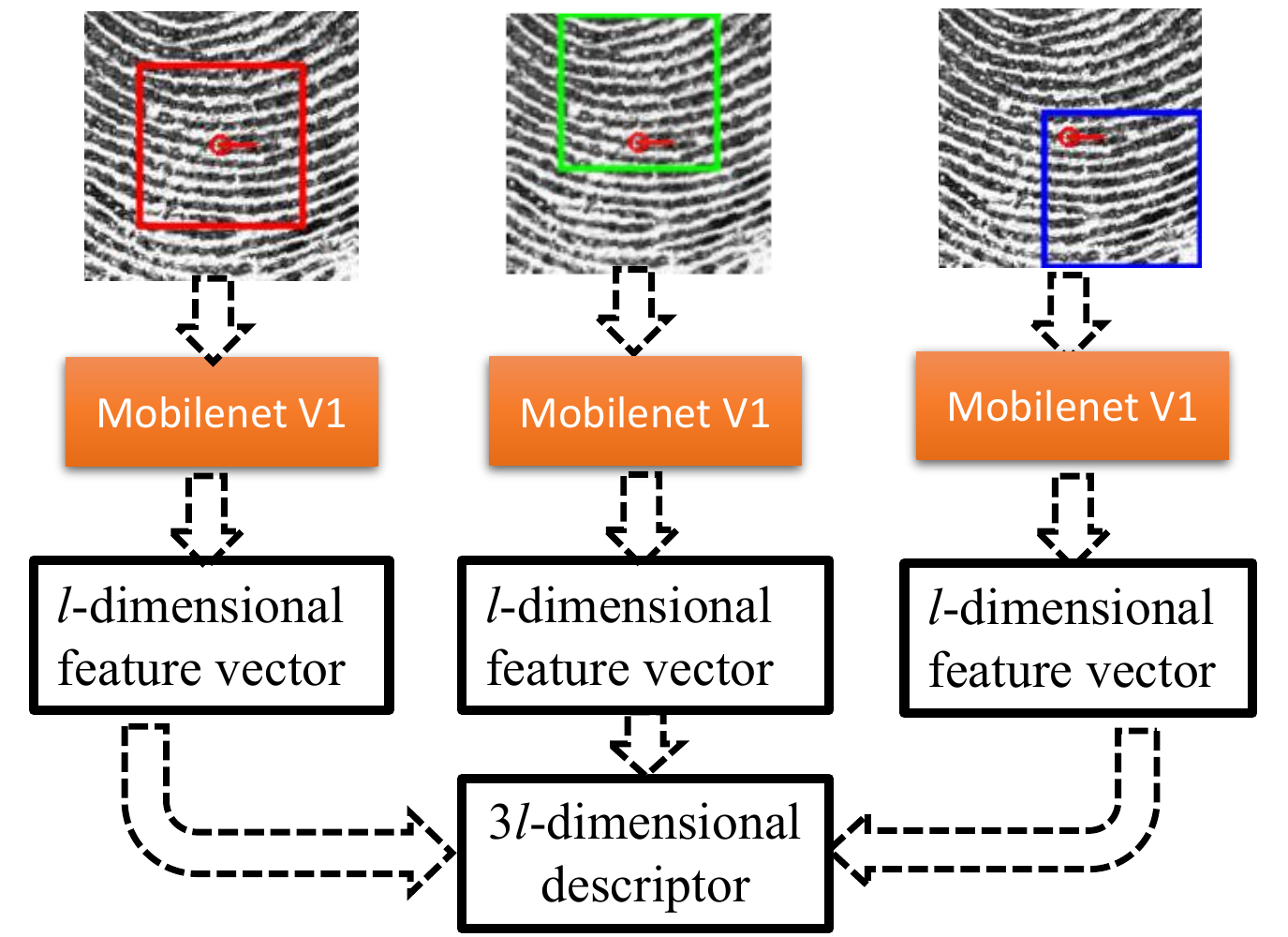}
			\caption{Extraction of minutia descriptor using CNN. } 	
	\label{fig:descriptor}
\end{figure}

A minutia descriptor contains attributes of the minutia based on the image characteristics in its neighborhood. Salient descriptors are needed to establish robust and accurate minutiae correspondences and compute the similarity between a latent and reference prints. Instead of specifying the descriptor in an ad hoc manner, Cao and Jain \cite{Cao2018PAMI} showed that  descriptors learned from local fingerprint patches provide better performance than ad hoc descriptors. Later they improved  both the distinctiveness  and the efficiency of descriptor extraction~\cite{Cao2018BTAS}.  Fig. \ref{fig:descriptor} illustrates the descriptor extraction process. The outputs ($l-$dimensional feature vector) of three patches around each minutia are concatenated to generate the final descriptor with dimensionality 3$l$.   Three values of $l$ ( i.e., $l$=32, 64, and 128), were investigated;  we empirically determine that  $l=64$ provides the best tradeoff between recognition accuracy and  computational efficiency. In this paper, we adopt the same descriptor as in ~\cite{Cao2018BTAS} , where the descriptor length  $ L=192$.

\begin{figure}[h]	
	\centering
	\includegraphics[width=0.95\linewidth]{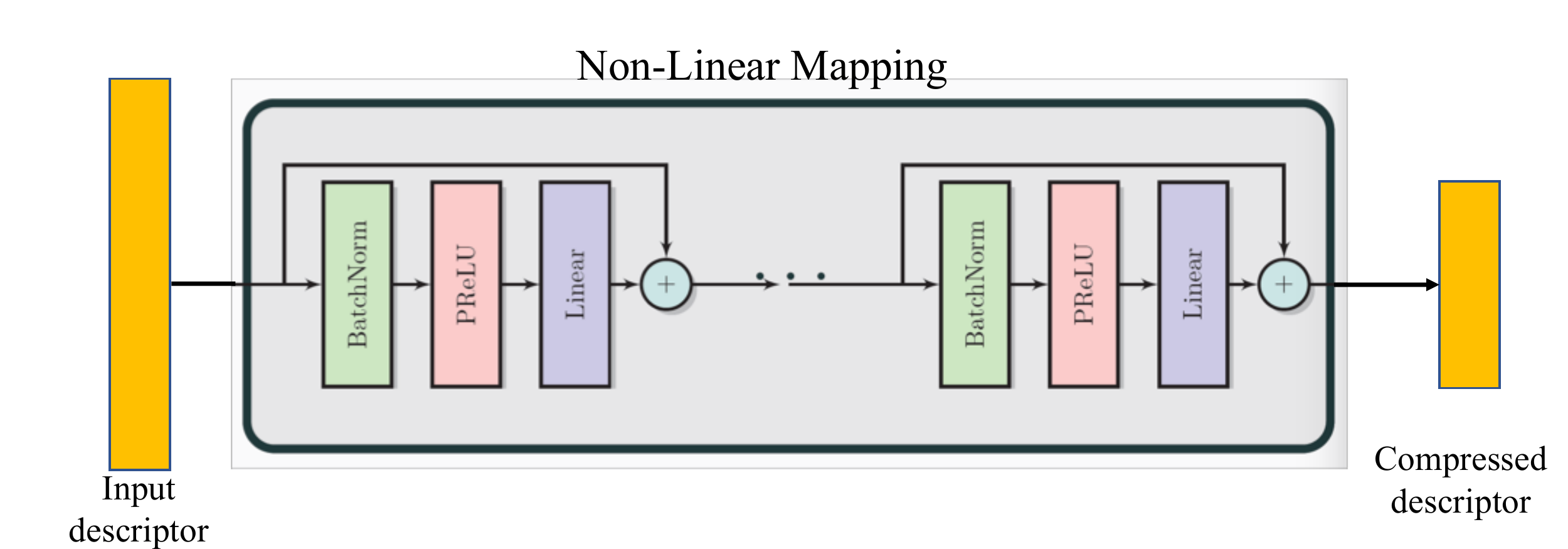}
			\caption{Framework for descriptor length reduction~\cite{Gong2018} which reduces descriptor length  from 192 to 96.} 	
	\label{fig:dim_reduction}
\end{figure}

\begin{figure}[h]	
	\centering
	\includegraphics[width=0.9\linewidth]{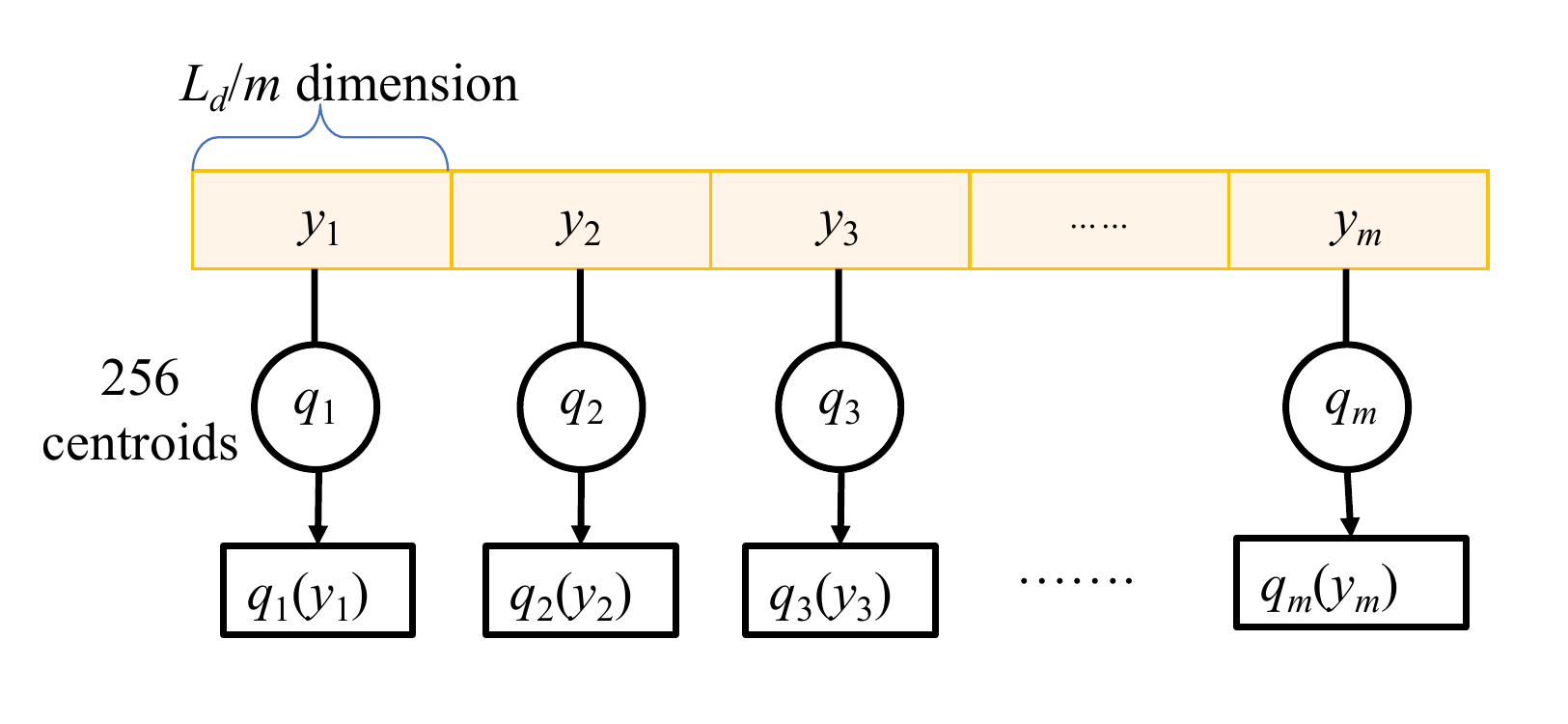}
			\caption{Illustration of descriptor product quantization. } 	
	\label{fig:product_quantization}
\end{figure}

Since there are a large number of virtual minutiae ($\sim 1,000$) in a texture template, further reduction of descriptor length is essential for improving the comparison speed between input latent and 100K reference prints. We utilized the non-linear mapping network of  Gong et al. \cite{Gong2018} for dimensionality reduction.  The network consists of four linear layers (see Fig. \ref{fig:dim_reduction}), where the objective is to minimize the distance between the cosine similarity of two input descriptors and  the corresponding cosine similarity of two output compressed descriptors. Empirical results show that the best value of the descriptor length in the compressed domain ($L_d$) in terms of recognition accuracy is 96.  In order to further reduce the virtual minutiae descriptor length, product quantization is adopted. Given a $L_d$-dimensional descriptor $y$, it is divided into $m$ subvectors, i.e., $y=[y_1|y_2|...|y_m]$, where each subvector is of size $L_d/m$.  The quantizer $q$ contains $m$ subquantizers i.e.,  $q(y)\mapsto [q_1(y_1)|q_2(y_2)|...|q_m(y_m)]$, where each subquantizer quantizes the input subvector into the closest centroid out of the 256 centroids trained by k-means clustering. Fig. \ref{fig:product_quantization} illustrates the product quantization process.  The distance $D(x,q(y))$ between an input 96-dimensional descriptor $x$ and a quantized descriptor $q(y)$ is computed as 
\begin{equation}
\label{eq:distance}
D(x,q(y))=\sum_{i=1}^{m}||x_i- c_{q(y_i)}^i||,
\end{equation}
 where $x_i$ is the $i$th subvector of $x$,  $c_{q(y_i)}^i$ is the $q(y_i)$th centroid of the $i$th subvector and $||\cdot||$ is the Euclidean distance. The final dimensionality of the descriptor of rolled prints is $m=16$.

% descriptor dimension reduction

\section{Reference Template Extraction}
Given that the quality of reference fingerprints, on average, is significantly better than latents, a smaller number of templates suffice for reference prints compared to latents.  Each reference fingerprint template consists of one minutiae template and one texture template. The model \emph{MinuNet\_reference} was used for minutiae detection on reference fingerprints. Since the reference fingerprint images were directly used for training, no preprocessing on the  reference fingerprint images is needed.  Fig. \ref{fig:minutiae_eg} show some examples of minutiae sets extracted on low quality and high quality rolled fingerprint images. For each minutia,  the descriptor is extracted following the approach shown in Fig. \ref{fig:descriptor} with descriptor length reduction via nonlinear mapping in Fig. \ref{fig:dim_reduction}. 
\begin{figure}[t]
	\centering
	\subfigure[]{
		\includegraphics[clip, trim=3cm 1cm 3cm 1cm, width=0.4\linewidth]{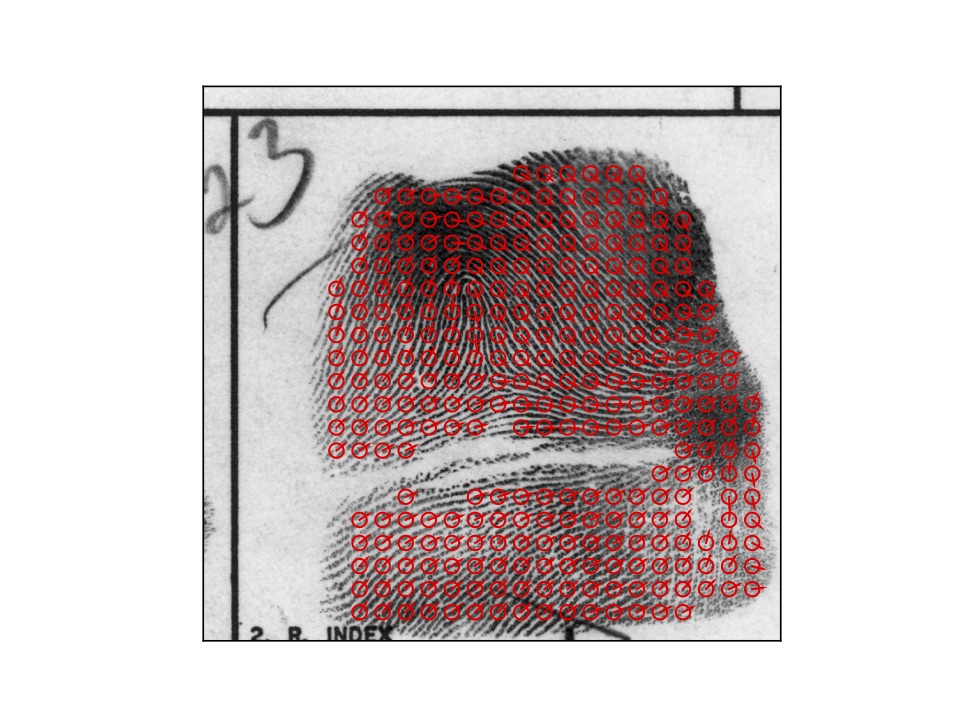}
	}\hspace{0.1cm}
	\subfigure[]{
		\includegraphics[clip, trim=3cm 1cm 3cm 1cm, width=0.4\linewidth]{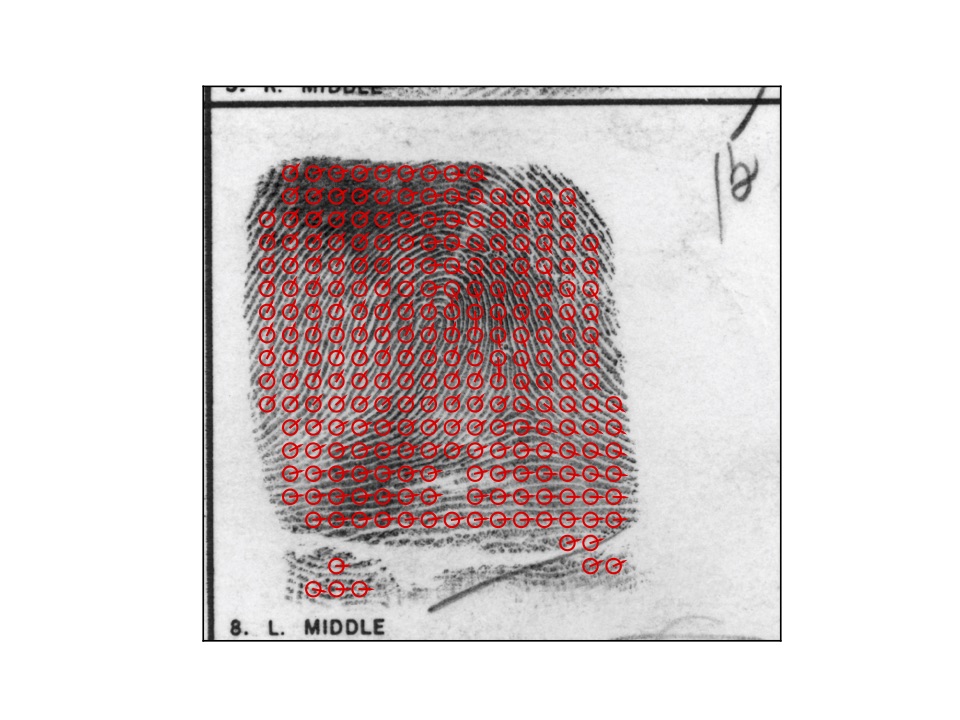}
	}           
	\caption{Virtual minutiae in  two rolled prints; stride size $s$=32.  }
	\label{fig:rolled_virtual}
	\vspace*{-5pt}
\end{figure}

\begin{comment}
\begin{figure}[h!]
	\centering
		\subfigure[]{
		\includegraphics[clip, width=0.425\linewidth]{019.jpg}
	}  \hspace{0.1cm}
	\subfigure[][]{
		\includegraphics[clip, trim=14cm 3cm 14cm 3cm, width=0.4\linewidth]{019_STFT.png}
	} \hspace{0.1cm}
	\subfigure[]{
		\includegraphics[clip, trim=14cm 3cm 14cm 3cm, width=0.4\linewidth]{019_constrast_STFT.png}
	}    \hspace{0.1cm}  
	\subfigure[]{
		\includegraphics[clip, trim=14cm 3cm 14cm 3cm, width=0.4\linewidth]{019_texture_Gabor.png}
	}    \hspace{0.1cm}  
		\subfigure[]{
		\includegraphics[clip, trim=14cm 3cm 14cm 3cm, width=0.4\linewidth]{019_constrast_Gabor.png}
	}   \hspace{0.1cm}
	\subfigure[][]{
		\includegraphics[clip, trim=14cm 3cm 14cm 3cm, width=0.4\linewidth]{019_common.png}
	}
	  \subfigure[][]{
		\includegraphics[clip, trim=14cm 3cm 14cm 3cm, width=0.6\linewidth]{019_three_minu_set.png}
	}
	\caption{Latent minutiae extraction. (a) Input latent, (b)-(e) minutiae sets after  i) STFT based enhancement, ii) contrast based enhancement, followed by STFT based enhancement, iii) decomposition followed by Gabor filtering,  and  iv) contrast based enhancement followed by Gabor filtering, respectively, and (g) common minutiae from (b)-(f).} %(g) Selected three minutiae sets (shown in (b), (d) and (f)) overlaid on input latent.}
	\label{fig:latent_minutiae_eg}
	\vspace*{-5pt}
\end{figure}
\end{comment}

\begin{figure}[h!]
	\centering
		\subfigure[]{
		\includegraphics[clip, width=0.4\linewidth]{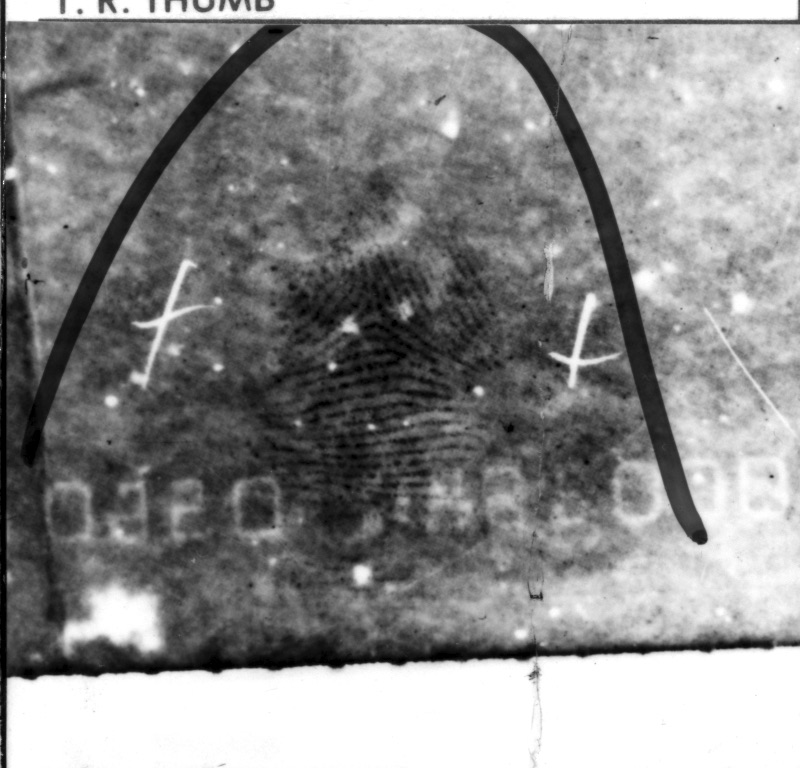}
	}  \hspace{0.1cm}
	\subfigure[][]{
		\includegraphics[clip, trim=3.5cm 1.4cm 3cm 1.4cm, width=0.4\linewidth]{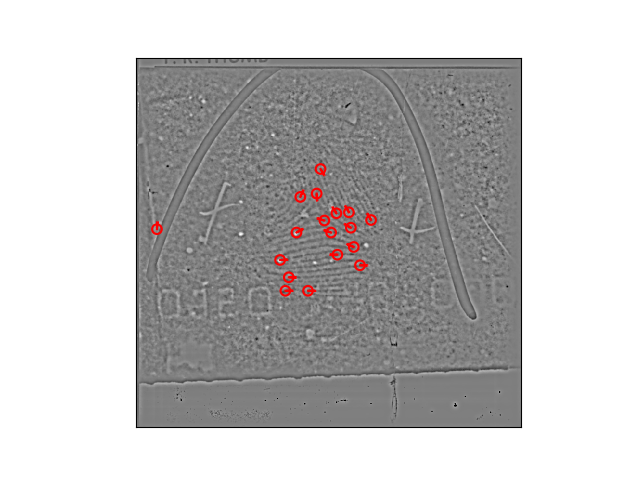}
	} \hspace{0.1cm}
	\subfigure[]{
		\includegraphics[clip, trim=3.5cm 1.4cm 3cm 1.4cm, width=0.4\linewidth]{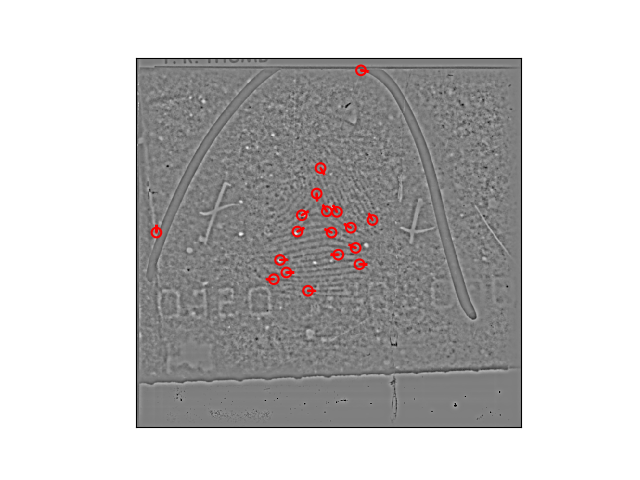}
	}    \hspace{0.1cm}  
	\subfigure[]{
		\includegraphics[clip, trim=3.5cm 1.4cm 3cm 1.4cm, width=0.4\linewidth]{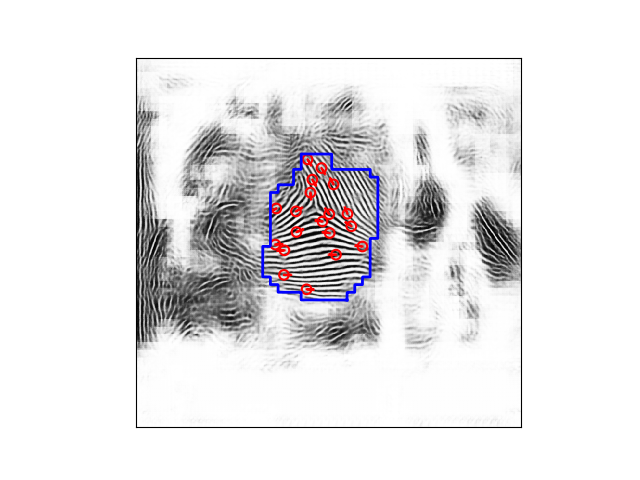}
	}    \hspace{0.1cm}  
		\subfigure[]{
		\includegraphics[clip, trim=3.5cm 1.4cm 3cm 1.4cm, width=0.4\linewidth]{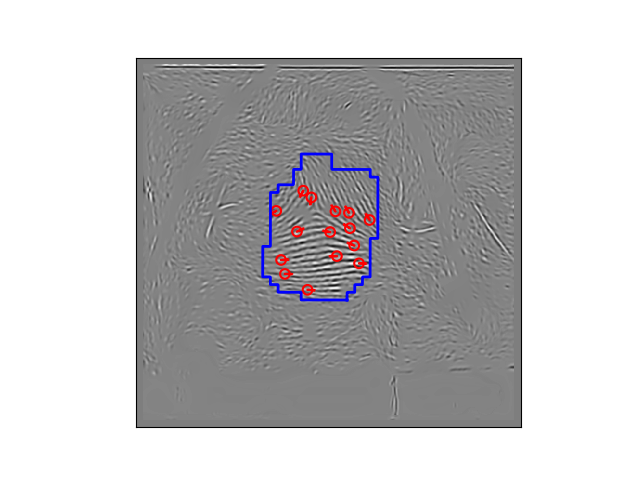}
	}   \hspace{0.1cm}
	\subfigure[][]{
		\includegraphics[clip, trim=3.5cm 1.4cm 3cm 1.4cm, width=0.4\linewidth]{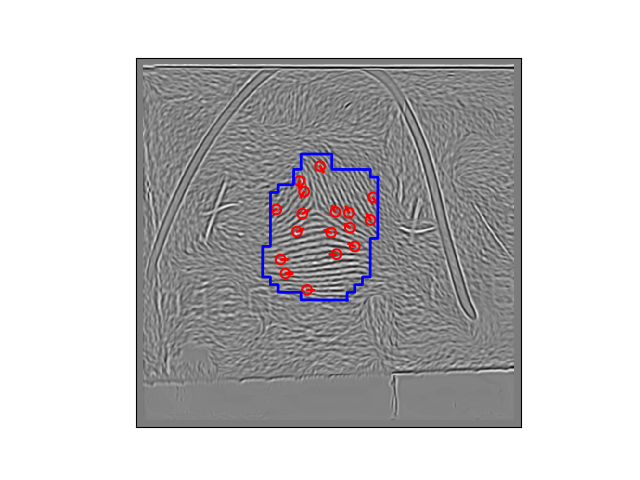}
	}
	  \subfigure[][]{
		\includegraphics[clip, trim=3.5cm 1.4cm 3cm 1.4cm, width=0.4\linewidth]{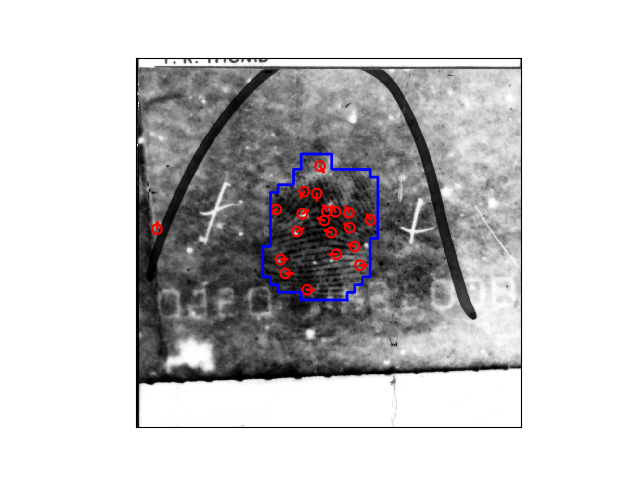}
	}
	\caption{Latent minutiae extraction. (a) Input latent, (b)-(f) automated extracted minutiae sets after  i) STFT based enhancement, ii) autoencoder based enhancement, iii) contrast based enhancement, followed by STFT based enhancement, iv) decomposition followed by Gabor filtering,  and  v) contrast based enhancement followed by Gabor filtering, respectively, and (g) common minutiae generated from (b)-(e) using majority voting.   Minutiae sets in (b), (d) and (g) are selected for matching.  Note that minutiae sets in (b) and (c) are extracted by \emph{MinuNet\_Latent}  but the mask is not used to remove spurious minutiae in case mask is inaccurate. } %In addition, manually marked minutiae are shown in red for a comparison.   (shown in blue) 
 	\label{fig:latent_minutiae_eg}
	\vspace*{-5pt}
\end{figure}

A texture template for reference prints is introduced in the same manner as for latents. The ROI for reference prints is defined by the magnitude of the gradient and the orientation field with a block size of $16\times 16$ pixels as in \cite{Chikkerur2007198}. %Fig. \ref{fig:reference_print} (a) and (b) shows the orientation fields within the ROIs of two example reference prints. 
The locations of virtual minutiae are sampled by raster scan with a stride of $s$ and their orientations are the same as the orientations of its nearest block in the orientation field. The virtual minutiae close to the mask border are ignored.  Fig. \ref{fig:rolled_virtual} shows  virtual minutiae extracted in two rolled prints. Similar to real minutiae, a  96-dimensional descriptor is first obtained using Fig. \ref{fig:descriptor}  and   Fig. \ref{fig:dim_reduction}, and then further reduced to $16$ dimensions using product quantization.

\section{Latent Template Extraction}

\begin{algorithm}
	\caption{Latent template extraction algorithm}\label{alg:latent_template}
	\begin{algorithmic}[1]
		%\Procedure{}{}
		\State \textbf{Input:} Latent fingerprint image
		\State \textbf{Output:} 3 minutiae templates and 1 texture template
		\State 	Enhance latent by autoencoder; estimate ROI, ridge flow and ridge spacing\;		
		\State	Process friction ridges: (i) STFT, (ii) contrast enhancement + STFT, (iii) autoencoder, (iv) decomposition + Gabor filtering and (v) contrast enhancement + Gabor filtering\;
		\State       Apply minutiae model \emph{MinuNet\_Latent}  to processed images (i) and (ii) in step 4 to generate minutiae sets 1 (Fig. \ref{fig:latent_minutiae_eg} (b)) and 2 (Fig. \ref{fig:latent_minutiae_eg} (c))\;		
		\State       Apply minutiae model \emph{MinuNet\_reference}  to  processed images (iii) - (v) in step 5 to generate minutiae sets 3 (Fig. \ref{fig:latent_minutiae_eg} (d)), 4 (Fig. \ref{fig:latent_minutiae_eg} (e)) and 5 (Fig. \ref{fig:latent_minutiae_eg} (f)) \;
		\State      Generate a common minutiae set 6 (Fig. \ref{fig:latent_minutiae_eg} (g)) using minutiae sets 1-5\;
		\State      Extract descriptors for minutiae sets 1, 3 and 6 to obtain the final 3 minutiae templates\;
		\State 	Generate a texture template using virtual minutiae and the associated descriptor\;
	%	\EndProcedure
	\end{algorithmic}
\end{algorithm}
 In order to extract complementary minutiae sets for latents, we apply two minutiae detection models, i.e.,  \emph{MinuNet\_Latent } and \emph{MinuNet\_reference}, to four differently processed latent images as described earlier. This results in five minutiae sets.  A common minutiae set (minutiae set 6) is obtained from these five minutiae sets using majority voting. A minutia is regarded as a common minutia if two out of the four minutiae sets contain that minutia, which means the distance between two minutiae locations is less than 8 pixels and the difference in  minutia orientation is less than $\pi/6$. Fig. \ref{fig:latent_minutiae_eg} shows these five minutiae sets.  For computational efficiency,  only minutiae sets 1, 3 and 6 are retained for matching.  Each selected minutiae set as well as the set of associated  descriptors form a minutiae template.  The texture template consists of the virtual minutiae located using ROI and ridge flow \cite{Cao2018BTAS}, and their associated descriptors.   \textbf{Algorithm} \ref{alg:latent_template} summarizes the latent template extraction process.

\section{Latent-to-Reference Print Comparison}
\begin{figure}[t]
	\centering
	\includegraphics[width=0.8\linewidth]{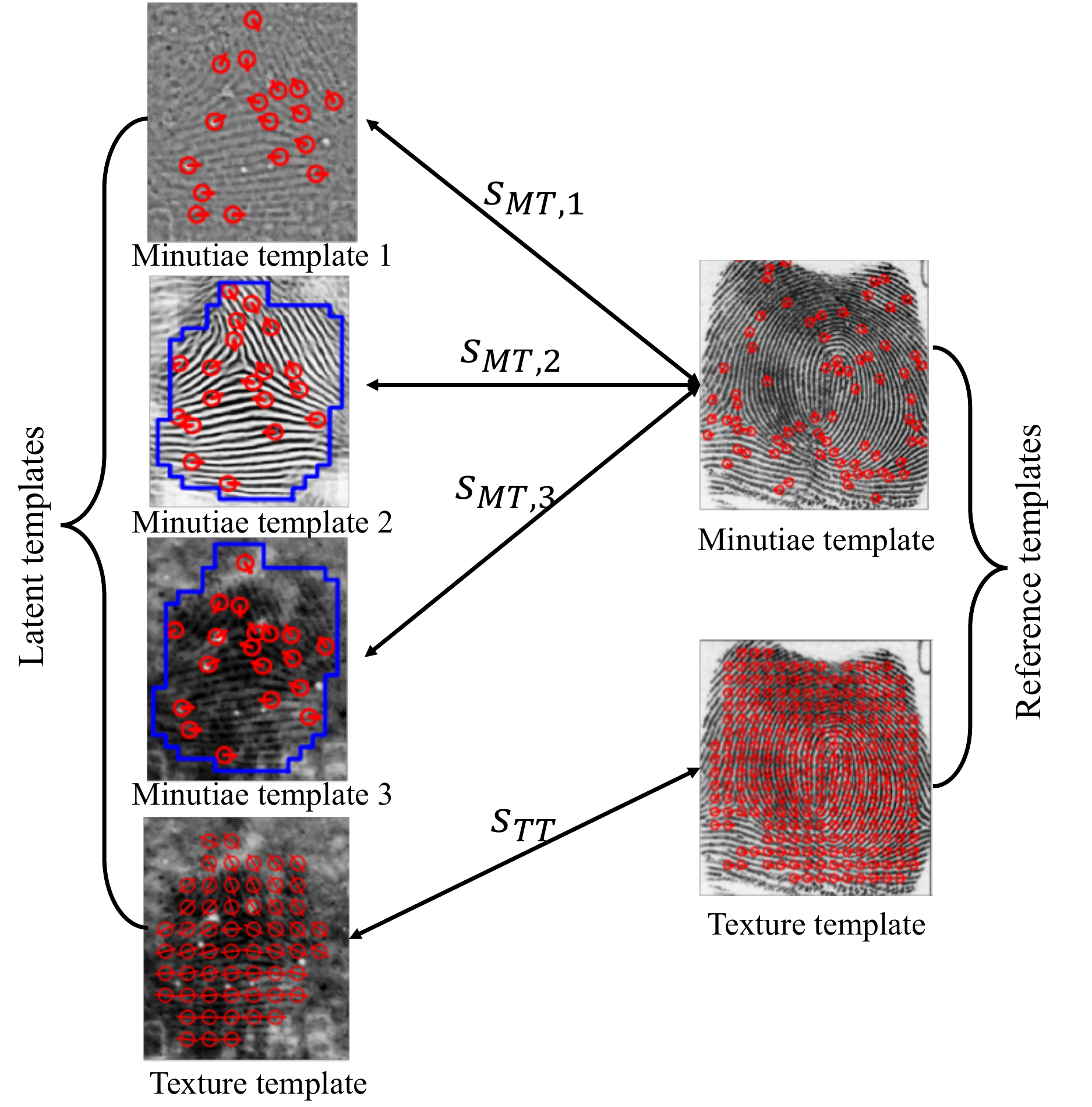}       
	\caption{ Latent-to-reference print templates comparison. Three latent minutiae templates are compared to one reference minutiae template, and the latent texture template is compared to the reference texture template.  Four comparison scores are fused to generate the final comparison score. }
	\label{fig:strategy}
	\vspace*{-5pt}
\end{figure}

Two comparison algorithms, i.e., minutiae template comparison and texture template comparison, are proposed for latent-to-reference comparison (See Fig. \ref{fig:strategy}).

\subsection{Minutiae Template Comparison}
\label{sec:minutiae_comparison}
Each minutiae template contains a set of minutiae points, including their $x$, $y$-coordinates and orientations, and their associated descriptors.  Let $M^l = \{m_i^l =(x_i^l,y_i^l,\alpha_i^l, d_i^l)\}_{i=1}^{n_l} $ denote a latent minutiae set with $n_l$ minutiae, where $(x_i^l,y_i^l)$, 
$\alpha_i^l$ and $d_i^l$ are $x-$ and $y-$coordinates, orientation and descriptor vector of the $i$th minutia, respectively. Let  $M^r = \{m_j^r =(x_j^r, y_j^r,\alpha_j^r, d_j^r)\}_{j=1}^{n_r}$  denote a reference print minutiae set with $n_r$ minutiae,  where $(x_j^r,y_j^r)$, $\alpha_j^r$  and $d_j^r$ are their  $x-$ and $y-$coordinates, orientation and descriptor of the $j$th reference minutia, respectively.  The comparison algorithm in \cite{Cao2018BTAS} is adopted for minutiae template comparison. For completeness, we summarize the minutiae template comparison algorithm in  \textbf{Algorithm} \ref{alg:minutiae}.
\begin{algorithm}
	\caption{Minutiae template comparison algorithm}\label{alg:minutiae}
	\begin{algorithmic}[1]
		%\Procedure{}{}
		\State \textbf{Input:} Latent minutiae template $M^l$ with $n_l$ minutiae and reference minutiae template $M^r$ with $n_r$ minutiae
		\State \textbf{Output:} Similarity score
		\State 	Compute the $n_l\times n_r$  similarity matrix ($S$) using the cosine similarity between descriptors\;		
		\State	Normalize the similarity matrix from $S$ to $S'$ using the approach in \cite{Feng2008PR}\;
		\State       Select the top $N$ ($N$=120) minutiae correspondences based on the normalized similarity matrix\;
		\State 	Remove false minutiae correspondences using simplified second-order graph matching\;
		\State	Remove additional false minutiae correspondences using full second-order graph matching\;
		\State       Compute similarity $s_{mt}$ between $M^l$ and $M^r$\;	
	%	\EndProcedure
	\end{algorithmic}
\end{algorithm}

\subsection{Texture Template Comparison}

Similar to the minutiae template, a texture template contains a set of virtual minutiae points, including their $x$, $y$-coordinates and orientations, and associated quantized descriptors. Let $T^l = \{m_i^l =(x_i^l,y_i^l,\alpha_i^l, d_i^l)\}_{i=1}^{n_l} $   and $T^r = \{m_j^r =(x_j^r, y_j^r,\alpha_j^r, d_j^r)\}_{j=1}^{n_r}$ denote a latent texture template and a reference texture template, respectively, where $d_i^l$ is a 96-dimensional descriptor of the $i$th latent minutia and $d_j^r$ is the $96/m$-dimensional quantized descriptor of the $j$th reference minutia. The overall texture template comparison algorithm is essentially the same as the  
 minutiae template comparison algorithm in  {\textbf{Algorithm}} \ref{alg:minutiae} with two main differences: i) descriptor similarity computation and ii) top $N$ virtual minutiae correspondences selection.  The similarity $s(d_i^l,d_j^r)$ between $d_i^l$ and  $d_j^r$ is computed as  $s(d_i^l,d_j^r) = D_0 -D(d_i^l,d_j^r)$, where $D_0$ is a threshold and $D(d_i^l,d_j^r)$ is defined in Eq. (\ref{eq:distance})  which can be computed offline. 

Instead of normalizing all scores and then selecting the top $N$ ($N=200$ for texture template comparison) initial virtual minutiae correspondences among all $n_l\times n_r$ possibilities, we select the top 2 reference virtual minutiae for each latent virtual minutiae based on virtual minutiae similarity and select the top $N$  initial virtual minutiae correspondences among $2\cdot n_l$ possibilities ($2\cdot n_l$ correspondences are all selected if $2\cdot n_l<=N$). In this way, we further reduce the computation time.

\subsection{Similarity Score Fusion}
Let $s_{MT,1}$, $s_{MT,2}$ and $s_{MT,3}$ denote  the similarities  between the three latent minutiae templates against the single reference minutiae template. Let $s_{TT}$ denote the similarity between the latent and reference texture templates.  The final similarity score $s$ between the latent and the reference print is computed as the weighted sum of $s_{MT,1}$, $s_{MT,2}$, $s_{MT,3}$ and $s_{TT}$ as below:
\begin{equation}
\label{eq:FinalScore}
s =\lambda_1 s_{MT,1} + \lambda_2 s_{MT,2} + \lambda_3 s_{MT,3}+\lambda_4 s_{TT},
\end{equation}
where $\lambda_1$, $\lambda_2$, $\lambda_3$ and  $\lambda_4$ are the weights that sum to 1; their values are empirically determined to be  1, $1$, $1$ and 0.3, respectively.

\subsection{Implementation}

\label{sec:implementation}

Both minutiae template comparison and texture template comparison algorithms are implemented in C++. In addition, matrix computation tool Eigen\footnote{https://github.com/libigl/eigen} is used for faster minutiae similarity computation.  OpenMP (Open Multi-Processing)\footnote{https://www.openmp.org/resources/openmp-compilers-tools/},  an application programming interface (API) that supports multi-platform shared memory multiprocessing programming, is used for code parallelization. Hence the latent-to-reference comparison algorithm can be executed on multiple cores simultaneously. The search speed  ($\sim$1.0 ms per latent to reference print comparison) on a 24-core machine   is able to achieve about 10-times speedup  compared to a single-core machine.

\section{Experiments} 
\begin{figure}[h]	
	\centering
	\includegraphics[width=0.95\linewidth]{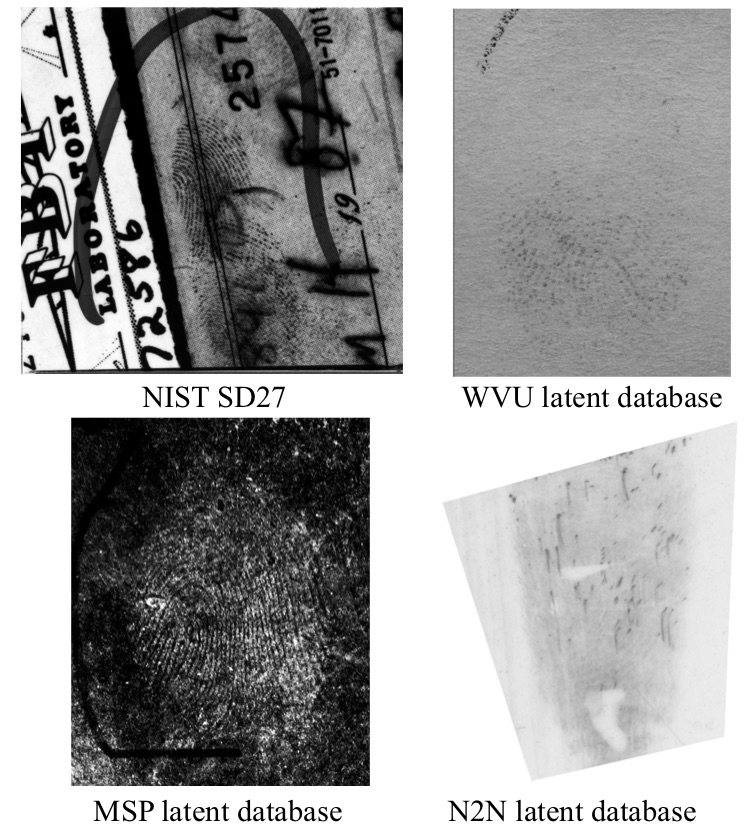}
			\caption{Examples of latents from the four databases. } 	
	\label{fig:latents}
\end{figure}

\label{sec:experiments}

%N2N database\footnote{https://www.nist.gov/programs-projects/n2n-fingerprint-capture-challenge},

In this report, three latent databases, NIST SD27 \cite{NISTDB27}, MSP  and WVU databases are used to evaluate the proposed end-to-end latent AFIS.  Table \ref{tab:databases} summarizes the three latent databases and Fig. \ref{fig:latents} shows some example latents. 
In addition to the mated reference
prints, we use additional reference fingerprints, from NIST SD14 \cite{NISTDB14} and a forensic agency,
to enlarge the reference database to 100,000 for search results reported here. We follow the protocol used in NIST ELFT-EFS \cite{Indovina2011}, \cite{Indovina2012} to evaluate the search performance of our system. 
%The performance is reported on two tasks: close-set identification and open-set identification. In the close-set setting, the query is always in the gallery while in the open-set setting, the query does not have to be in the gallery. 

%The NIST SD27 contains 258 latent fingerprints with their mated reference fingerprints. The MSP latent database contains 2,074 latent fingerprint images. Since some latent fingerprint images were used for training minutiae extractor and the resolution of some latents are much higher than 500  dpi, 1,200 latents are finally used for search experiments. The N2N latent database contains 10,000 latents with their mated reference prints from 200 subjects.  Note that the NIST SD27 and MSP latent databases are collections of latents from the casework of forensics agencies, whereas N2N latent database was collected in  a laboratory setting. As such, the characteristics of these three databases are quite different in terms of background noise, ridge clarity, and the number of minutiae. See Fig. \ref{fig:latents} for a comparison of the images in the three latent databases.

\begin{table}[t]
\caption{Summary of  latent databases.}
\vspace*{-10pt}
\begin{center}
\centering
\begin{tabular}{|p{2cm}|p{2cm}|p{2.5cm}|}
\hline
Database & No. of latents & Source \\
\hline
\hline
NIST SD27&  258 &         Forensic agency  \\
\hline
MSP        &  1,200 &         Forensic agency \\
\hline
WVU        &   449 &         Laboratory \\
\hline
N2N         &  10,000 &      Laboratory\\
\hline
\end{tabular}
\end{center}
\label{tab:databases}
\vspace*{-10pt}
\end{table}%

\subsection{Evaluation of Descriptor Dimension Reduction}
We evaluate the non-linear mapping based descriptor dimension reduction and product quantization on NIST SD27 against a 10K gallery.  Non-linear mapping is adopted to reduce the descriptor length of both real minutiae and virtual minutiae. Three different descriptor lengths, i.e., 128, 96 and 64, are evaluated.   Table \ref{tab:reduction} compares the search performance of different descriptor lengths. There is a slightly drop for 96- and 48-dimensional descriptors, but a significantly drop for 48-dimensional descriptors. 

Because of the large number of virtual minutiae, we further reduce the descriptor length of virtual minutiae using product quantization.  Table \ref{tab:PQ} compares the search performance of texture template on NIST SD27 using three different number of subvectors of 96-dimensional descriptors, i.e., $m=24, 16$ and $12$.  $m=16$ achieves a good tradeoff between accuracy and feature length.  Hence, we use non-linear mapping to reduce  the descriptor length from 192 dimension to  96 dimension and then further reduce virtual minutiae descriptor length to $m=16$ using product quantization in the following experiments.

\begin{table}[t]
\caption{Search performance on NIST SD27 after non-linear mapping}
\vspace*{-10pt}
\begin{center}
\centering
\begin{tabular}{|p{2cm}|p{1.8cm}|p{1.8cm}|p{1.8cm}|}
\hline
Dimension & Rank-1 & Rank-5 & Rank-10\\
\hline
\hline
192  &    72.5\%   &   77.5\%     & 79.5\%  \\
%\hline
%128      &    71.8\%   &  77.5\%        &  79.3\% \\
\hline
96       &    71.3\%   &   77.5\%      & 79.1\% \\
\hline
48      &     61.6\%  &     67.8\%    & 70.9\%\\
\hline
\end{tabular}
\end{center}
\label{tab:reduction}
\vspace*{-10pt}
\end{table}%

\begin{table}[t]
\caption{Search performance of texture template on NIST SD27 using different product quantization (PQ) settings.}
\vspace*{-10pt}
\begin{center}
\centering
\begin{tabular}{|p{2cm}|p{1.8cm}|p{1.8cm}|p{1.8cm}|}
\hline
Value of $m$ & Rank-1 & Rank-5 & Rank-10\\
\hline
\hline
Without PQ  &    65.5   &  70.5\%        &  74.8\%\\
\hline
$m=$24  &    64.3\%   &      69.8\%   &  72.1\%  \\
\hline
$m=$16      & 63.6\%      & 69.4\%        & 71.3\% \\
\hline
$m=$12       &  58.9\%    &     65.1\%    & 69.8\% \\
\hline
%12       &       &         & \\
% \hline
\end{tabular}
\end{center}
\label{tab:PQ}
\vspace*{-10pt}
\end{table}%

\begin{figure*}[h!]
	\centering
		\subfigure[]{
		\includegraphics[clip, trim=0cm 0cm 0cm 0cm, width=0.41\linewidth]{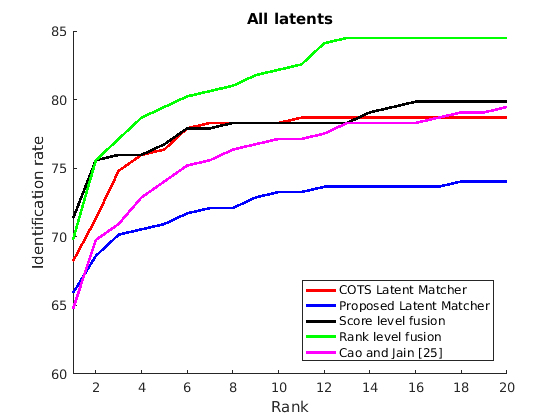}
	}  \hspace{0.5cm}
	\subfigure[][]{
		\includegraphics[clip, trim=0cm 0cm 0cm 0cm, width=0.41\linewidth]{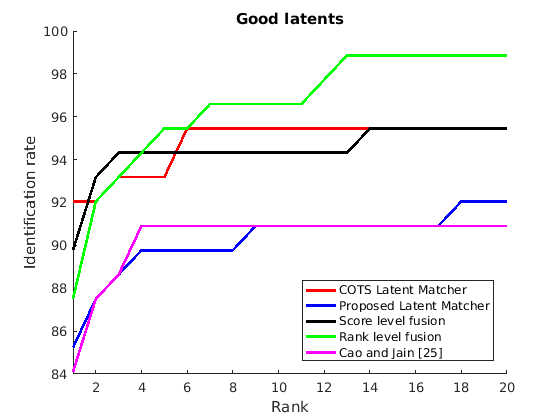}
	}
	\subfigure[]{
		\includegraphics[clip, trim=0cm 0cm 0cm 0cm, width=0.41\linewidth]{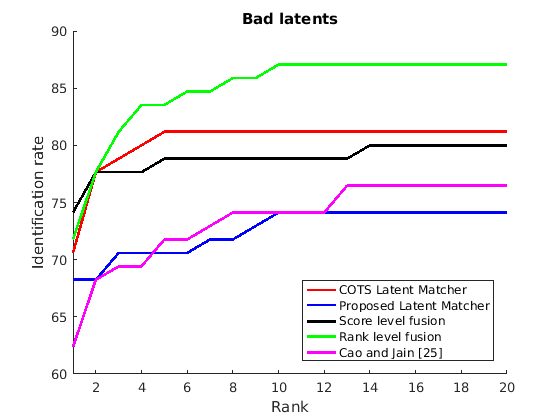}
	}   \hspace{0.5cm}
	\subfigure[]{
		\includegraphics[clip, trim=0cm 0cm 0cm 0cm, width=0.41\linewidth]{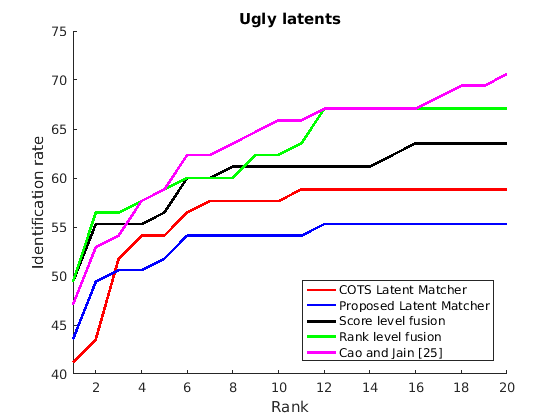}
	}    
	\caption{Cumulative Match Characteristic (CMC) curves of our latent search system and COTS latent AFIS, their score-level and rank-level fusions, and semi-automatic algorithm of Cao and Jain \cite{Cao2018PAMI} on (a) all 258 latents in NIST SD27, (b) subset of 88 ``good'' latents, (c) subset of 85 ``bad'' latents and (d) subset of 85 ``ugly'' latents. 
Note that the scales of the y-axis in these four plots are different to accentuate the differences between the different curves. }
	\label{fig:NISTSD27}
	\vspace*{-5pt}
\end{figure*}

\begin{figure}[h!]	
	\centering
	\includegraphics[width=0.85\linewidth]{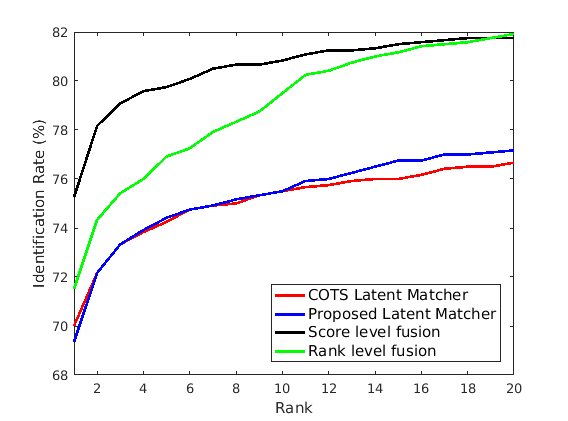}
			\caption{CMC curves of our latent search system, COTS latent AFIS, and score-level and rank-level fusion of the two systems on the MSP latent database against 100K reference prints. } 	
	\label{fig:MSP}
\end{figure}

\begin{figure}[t]
	\centering
		\subfigure[]{
		\includegraphics[clip, trim=2cm 1cm 2cm 1cm, width=0.45\linewidth]{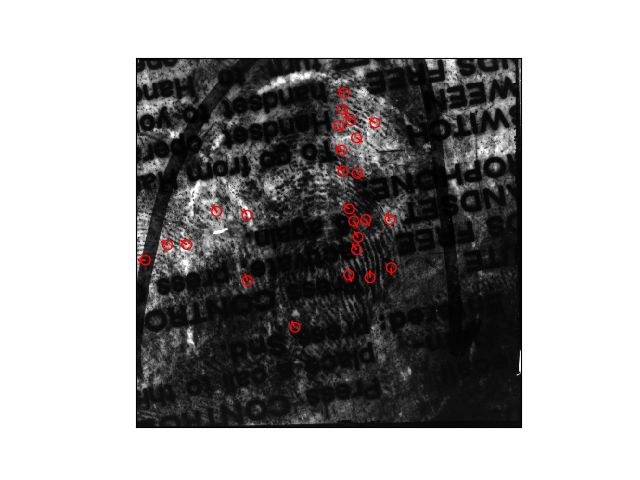}
	}  
	\subfigure[][]{
		\includegraphics[clip, trim=2cm 1cm 2cm 1cm, width=0.45\linewidth]{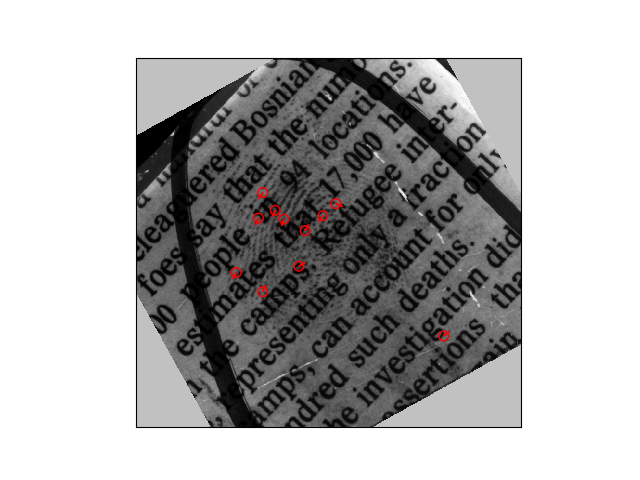}
	}
	\subfigure[]{
		\includegraphics[clip, trim=2cm 1cm 2cm 1cm, width=0.45\linewidth]{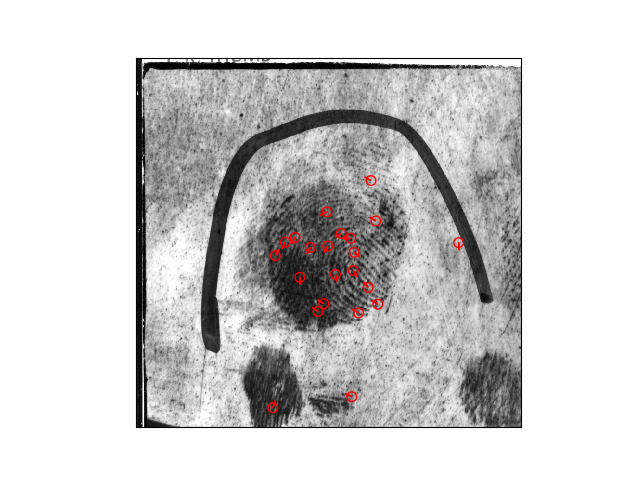}
	}     
	\subfigure[]{
		\includegraphics[clip, trim=2cm 1cm 2cm 1cm, width=0.45\linewidth]{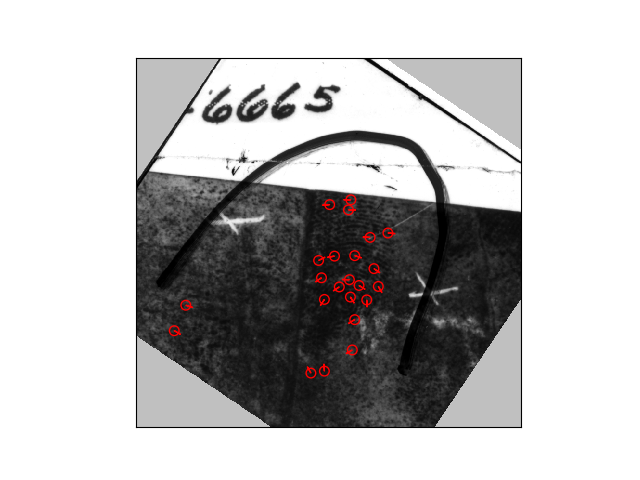}
	}       
	\caption{Our latent AFIS can retrieve the true mates of latents in (a) and (b) at rank-1 which the COTS latent AFIS cannot. COTS latent AFIS can retrieve the mates of latents in (c) and (d) at rank-1 while our latent AFIS cannot. One minutiae set extracted by our AFIS  is overlaid on each latent.  These latents are from the NIST SD27 database.}
	\label{fig:NISTSD27_eg}
	\vspace*{-5pt}
\end{figure}

\begin{figure}[h!]	
	\centering
	\includegraphics[width=0.85\linewidth]{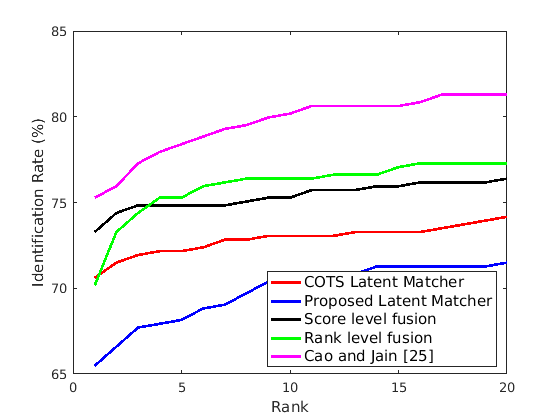}
			\caption{CMC curves of our latent search system,
COTS latent AFIS.  Score-level and rank-level fusions of
the two systems on the WVU latent database against 100K reference prints  show that the bot the fusion schemes boost the overall recognition accuracy significantly. } 	
	\label{fig:WVU}
\end{figure}

\begin{figure}[h!]	
	\centering
	\includegraphics[width=0.85\linewidth]{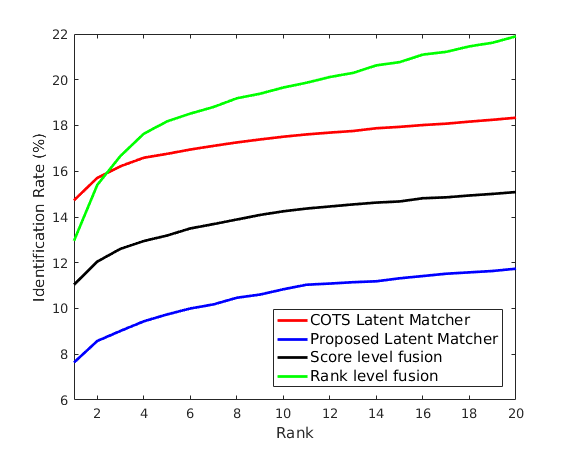}
			\caption{CMC curves of our latent search system, COTS latent AFIS, and score-level and rank-level fusion of the two systems on the N2N latent database against 100K reference prints. } 	
	\label{fig:N2N}
\end{figure}

\begin{figure}[t]
	\centering
		\subfigure[]{
		\includegraphics[clip,width=0.4\linewidth]{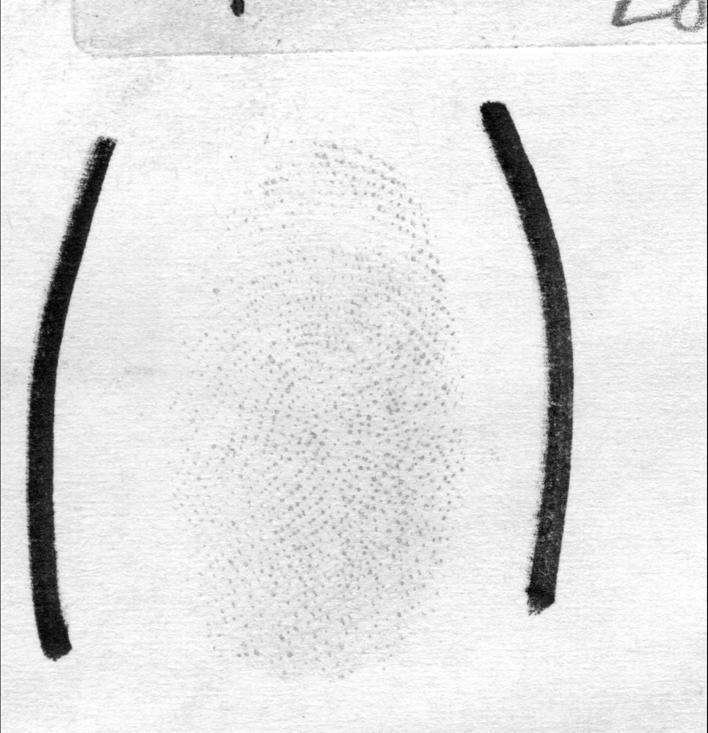}
	}  
	\subfigure[][]{
		\includegraphics[clip,width=0.4\linewidth]{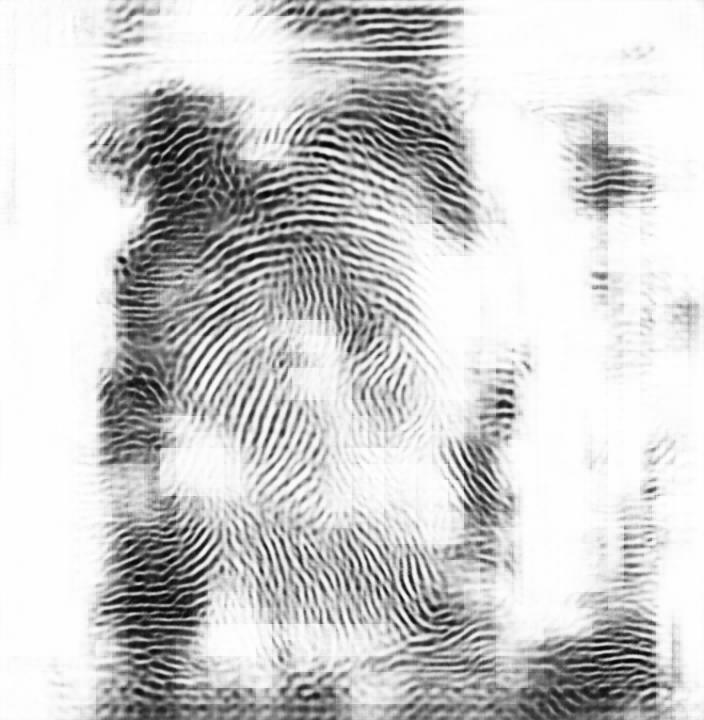}
	}   
	\caption{A failure case in the WVU latent database. Because the training database does not have any dry fingerprints like the latent image in (a), the enhanced latent image in (b) by the Autoencoder does not look good.  }
	\label{fig:WVU_eg}
	\vspace*{-5pt}
\end{figure}

\subsection{Search Performance}

We benchmark the proposed latent AFIS against one of the best COTS latent AFIS\footnote{The latent COTS used here is one of the top-three performers in the
NIST ELFT-EFS evaluations  \cite{Indovina2011}, \cite{Indovina2012} and the method in \cite{Cao2018PAMI}.  Because of our non-disclosure
agreement with the vendor, we cannot disclose its name.} as determined in NIST evaluations. Two fusion strategies, namely score-level fusion
(with equal weights) and rank-level fusion (top-200 candidate lists are fused using Borda count), are adopted to determine if the proposed algorithm and COTS latent AFIS have complementary search capabilities. In addition,  the algorithm proposed in \cite{Cao2018PAMI} is also included for comparison on NIST SD27 and WVU databases.  

 The performance is reported based on close-set identification where the query is assumed to be in the gallery. Cumulative Match Characteristic (CMC) curve is used for performance evaluation.  Fig. \ref{fig:NISTSD27} compares the five CMC curves on all 258 latents in NIST SD27 as well as subsets of latents of three different quality levels (good, bad and ugly) and  Fig. \ref{fig:MSP}  compares the four CMC curves on 1,200 latents in MSP latent database.  On both operational latent databases, the performance of our proposed latent AFIS is comparable to that of COTS latent AFIS. In addition, both rank-level and score-level fusion of two latent AFIS can significantly boost the performance, which indicates that these two AFIS provide complementary information. Figs.  \ref{fig:NISTSD27_eg} (a) and (b) show two examples that our latent AFIS can retrieve their true mates at rank-1 but the COTS AFIS cannot due to overlap between background characters and friction ridges. Figs. \ref{fig:NISTSD27_eg} (c) and (d) show two failure cases of the proposed latent AFIS due to the broken ridges. The rank-1 accuracy of proposed latent AFIS on NIST SD27  is slightly higher than the algorithm proposed in \cite{Cao2018PAMI} even though manually marked ROI was used in \cite{Cao2018PAMI}.  

% score level fusion is 5.25% higher than NEC
The five CMC curves on 449 latents  in WVU database are compared in  Fig. \ref{fig:WVU} and the four CMC curves on 10,000 latents in N2N database are compared in  Fig. \ref{fig:N2N}.  Both WVU and N2N databases were collected in laboratory.  The latents in these two latent databases are dry (ridges are broken), and are significantly from operational latents which were used for fine-tuning minutiae detection model and rolled prints which were used for training  Autoencoder for enhancement,  the minutiae detection model and enhancement model do not work well on WVU latent database. This explains why the performance of the proposed latent AFIS is lower than COTS latent AFIS. Fig. \ref{fig:WVU_eg} shows some examples where the enhancement model fails.  This indicates that additional dry fingerprints are needed for proposed training for deep learning based approaches.

 %rank-1 the proposed latent AFIS is be inferior to, 
% as the performance of  \cite{Cao2018PAMI} is not available. The performance of our end-to-end latent recognition system is comparable to that of the COTS latent AFIS on both WVU and MSP database although the rank-1 accuracy of our latent AFIS is 3.1\% lower than the COTS latent AFIS on NIST SD27. Fusion of these two latent AFIS, especially rank-level fusion, can significantly boost the overall performance on all three latent databases. However, our overall performance is inferior to \cite{Cao2018PAMI} because manual cropping was used in  \cite{Cao2018PAMI}. 
% The rank-1 identification accuracy of proposed latent AFIS is 1.3\% higher than that in \cite{Cao2018PAMI}.

% \subsection{Close-set Identification}
\label{sec:close-set }

\section{Summary}
We present the design and prototype of an  end-to-end fully automated latent search system and benchmark its performance
against a leading COTS latent AFIS. The contributions of this paper are as follows:
\begin{itemize}
\item Design and prototype  of the first fully automated end-to-end latent search system different curves.  
\item Autoencoder-based latent enhancement and minutiae detection. 
\item Efficient latent-to-reference print comparison.  One latent search against 100K reference prints can be completed in 100 seconds on a machine with Intel(R) Xeon(R) CPU E5-2680 v3@2.50GHz.
\end{itemize}

There are still a number of challenges we are trying to address listed below.
\begin{itemize}
\item  Improvement in automated cropping module. The current cropping algorithm does not perform well on dry latents in WVU and N2N databases.
\item  Obtain additional operational latent databases for robust training for various modules in the search system.
\item Include additional features, e.g., ridge flow and ridge spacing, for similarity measure.
\end{itemize}

% or


\ifCLASSOPTIONcompsoc
  % The Biometrics Council usually uses the plural form
  \section*{Acknowledgments}
\else
  \section*{Acknowledgment}
\fi

This research is based upon work supported in part by the
Office of the Director of National Intelligence (ODNI), Intelligence
Advanced Research Projects Activity (IARPA),
via IARPA R\&D Contract No. 2018-18012900001. The
views and conclusions contained herein are those of the
authors and should not be interpreted as necessarily representing
the official policies, either expressed or implied,
of ODNI, IARPA, or the U.S. Government. The U.S. Government
is authorized to reproduce and distribute reprints
for governmental purposes notwithstanding any copyright
annotation therein.

\ifCLASSOPTIONcaptionsoff
  \newpage
\fi



%\bibliographystyle{IEEEtran}

{\small
\bibliographystyle{IEEEtran}
\bibliography{FM_ref}


%\end{IEEEbiography}


%\newpage



%\vfill


\end{document}